\documentclass[twoside]{article}

\usepackage[preprint]{aistats2027}
\usepackage{titletoc}

\newcommand{\R}{\mathbb R}
\newcommand{\E}{\mathbb E}
\newcommand{\M}{\mathcal M}
\newcommand{\Z}{\mathcal Z}
\newcommand{\pdata}{p_{\mathrm{data}}}
\newcommand{\tr}{\operatorname{tr}}
\newcommand{\Cov}{\operatorname{Cov}}
\newcommand{\supp}{\operatorname{supp}}
\newcommand{\dist}{\operatorname{dist}}
\newcommand{\argmin}{\operatorname*{arg\,min}}

\newcommand{\post}{\mathrm{post}}

\usepackage[round]{natbib}

\usepackage{amsmath,amssymb,amsthm,mathtools}

\usepackage{graphicx}
\usepackage{booktabs,tabularx,array,multirow}
\usepackage{microtype}
\usepackage{todonotes}
\usepackage{placeins}
\usepackage{hyperref}
\usepackage{url}
\usepackage[table,dvipsnames]{xcolor}

\usepackage[most]{tcolorbox}
\newtcolorbox{takehomebox}[1][]{%
  enhanced, breakable,
  colback=teal!6, colframe=teal!65!black,
  boxrule=0pt, leftrule=2.5pt, sharp corners,
  left=6pt, right=6pt, top=4pt, bottom=4pt,
  before skip=8pt, after skip=8pt,
  fontupper=\small,
  #1}

\begin{document}

%

%

\runningauthor{Tanja Bien, Samir Darouich, Lluís Pastor-Pérez, Vinh Tong, Loay Mualem, Mathias Niepert}
\runningtitle{Characterizing and Mitigating Distributional Bias in Physics-Informed Generative Models}

\twocolumn[

\aistatstitle{Characterizing and Mitigating Distributional Bias\\
in Physics-Informed Generative Models}

\aistatsauthor{
    Tanja Bien \\ {\normalfont University of Stuttgart}
    \And
    Samir Darouich \\ {\normalfont University of Stuttgart}
    \And
    Lluís Pastor-Pérez \\ {\normalfont University of Stuttgart}
    \AND
    Vinh Tong \\ {\normalfont University of Stuttgart}
    \And
    Loay Mualem \\ {\normalfont University of Stuttgart}
    \And
    Mathias Niepert \\ {\normalfont University of Stuttgart}
}

\aistatsaddress{}
]
\begin{abstract}
Physics-informed generative models augment a data-fitting objective with residual penalties that encourage physical constraints to hold in generated samples. However, improving constraint satisfaction need not improve fidelity to the data distribution. We study this conflict in residual-regularized diffusion models and identify two possible consequences. First, the same prediction error can incur different residual costs near different feasible states, preferring states with lower residual sensitivity (avoidance). Second, this preference can extend to feasible states outside the data support, where the residual also vanishes (leakage). Motivated by this analysis, we propose a data-anchored penalty that weights prediction errors by the residual Jacobian at the clean training sample and normalizes the resulting magnitude. This removes the incentive to relocate to less sensitive states without changing the model architecture or sampling procedure. Experiments on a controlled synthetic system, three PDE systems, and molecular transition-state generation with flow matching demonstrate the benefits of anchoring. Code is available at \url{https://github.com/TanjaBien/anchored_diffusion}.
\end{abstract}

\section{Introduction}
\label{sec:intro}

Diffusion and flow matching models learn vector fields that guide the transformation of samples from a simple reference distribution into samples resembling the training data. They are increasingly applied in scientific machine learning
\citep{bose2024se,srivastava2024precipitation,bastek2025physics,
tong2025raoblackwell,darouich2026curvature}, where generated samples
are often expected to satisfy known physical constraints. These may be expressed through governing partial differential equations (PDEs), as in physical field generation, or through force-based conditions, as in molecular transition-state generation. Learning from physically valid examples, however, does not guarantee that generated samples satisfy these constraints.

To encourage constraint satisfaction, physics-informed generative models add a physical residual penalty to the training loss \citep{bastek2025physics, baldan2026physics}. A residual measures how much a prediction violates a constraint and vanishes when the constraint is satisfied. The additional loss is intended to improve physical accuracy while retaining the distribution learned from data. However, a lower expected residual does not imply a better approximation of that distribution \citep{baldan2026physics, chang2026stabilizing}.

In this work, we examine how this conflict arises from the residual loss favoring some feasible states over others. The residual is evaluated at the diffusion model's prediction, and its sensitivity to prediction errors can vary across states. An error of the same magnitude and direction can therefore incur a large residual near one feasible state and a small residual near another, while the standard squared prediction error remains unchanged. Moving a prediction toward a less sensitive state can then reduce the combined loss even when it increases the distance to the clean training target. We call this preference \emph{avoidance}. Furthermore, physical constraints specify which states are admissible, but not which occur in the data or how frequently they occur. The residual can therefore vanish at states outside the data support, allowing the preference for less sensitive states to extend to states absent from the data support. We call this possibility \emph{leakage}.

We formalize these mechanisms by comparing predictions under the same perturbations applied to different feasible states. This gives an exact condition under which the reduction in residual loss compensates for the additional denoising error. For small perturbations, the comparison is governed by the residual Jacobian, which describes how the residual changes as the predicted state changes. The analysis applies to differentiable residual maps and does not require a PDE.

These observations motivate a data-anchored residual loss. Instead of evaluating the residual at the prediction, we use the Jacobian of the residual at the clean training sample to weight the prediction error. The Jacobian remains fixed as the prediction changes, removing the incentive to move toward a state with a less sensitive Jacobian. We normalize this weighting to limit the influence of large Jacobian magnitudes. The resulting objective requires no change to the model architecture or sampling procedure. We characterize its optima for individual training samples and after averaging over the set of possible clean targets given a noisy input. Finally, we evaluate residual accuracy and fidelity to the data on a controlled synthetic system and the Burgers, Darcy, and advection equations, as well as molecular transition-state generation with flow matching.

\section{Background and Setup}
\label{sec:setup}

\subsection{Partial differential equations}

Let $u(\xi,\tau)$ be a field defined at spatial position
$\xi\in\Omega\subset\R^p$ and physical time $\tau\in[0,T]$. Its governing
PDE, with differential operator $\mathcal F$, is
\begin{equation}
  \mathcal F[u](\xi,\tau)=0,\qquad \xi\in\Omega,\quad \tau\in[0,T].
  \label{eq:pde}
\end{equation}
The PDE generally admits a family of solutions, with initial and boundary conditions selecting a particular field. On a fixed grid, this field is represented by $x\in\R^d$. Each sample contains the complete discretized space--time trajectory $u(\xi,\tau)$, and $d$ is the total number of field values on this grid. The residual operator $R:\R^d\to\R^m$ measures violations of the discretized PDE, with $m$ residual equations \citep{xu2018spectral}. A discrete field satisfies the PDE when $R(x)=0$. 

\subsection{Physics-constrained diffusion models}
Diffusion models learn to
reverse a fixed corruption process \citep{ho2020denoising, songICLR2021scorebasedSDE}. Given a clean sample $x_0$, the forward process produces
\[
  x_t=\sqrt{\bar\alpha_t}\,x_0+
      \sqrt{1-\bar\alpha_t}\,\varepsilon,
  \, \varepsilon\sim\mathcal N(0,I),
  \,
  \bar\alpha_t=\prod_{s=1}^t(1-\beta_s).
\]
The variance schedule $\{\beta_s\}$ is chosen so that the terminal $x_t$
is approximately standard Gaussian. A network learns to recover the clean
sample from $x_t$.

Physics-constrained training \citep{bastek2025physics, rixner2021probabilistic}
adds the squared residual of the clean-sample prediction
$\hat x_0 := \hat x_0(x_t,t)$ to the denoising loss:
\begin{equation}
\begin{aligned}
  \mathcal L
  &= \E_{t,x_0,x_t}
  \left[
    \lambda_t\|x_0-\hat x_0\|^2
    + c_t\|R(\hat x_0)\|^2
  \right], \\
  &\text{with}\quad
  c_t := c\,\omega_t,
  \qquad
  \eta_t := \frac{c_t}{\lambda_t}.
\end{aligned}
\label{eq:objective}
\end{equation}

\begin{figure*}[t!]
\centering
\includegraphics[width=0.86\textwidth]{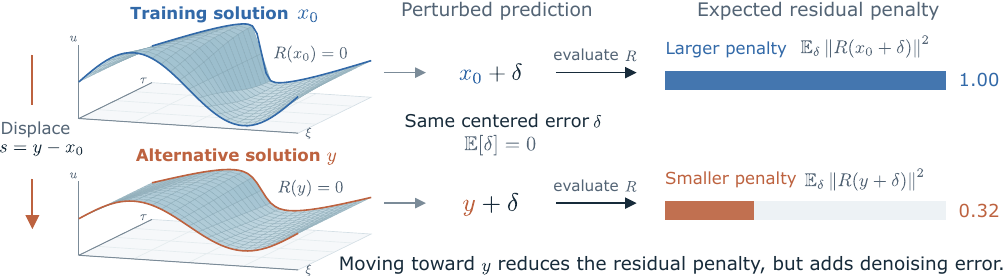}
\caption{Illustration of \emph{avoidance}. The training solution $x_0$ and an alternative solution $y$ both satisfy $R=0$, but the same zero-mean perturbation $\delta$ produces a smaller expected residual penalty near the less sensitive solution $y$. Shifting predictions from $x_0+\delta$ to $y+\delta$ therefore reduces the residual penalty while adding an expected denoising cost of $\|y-x_0\|^2$. The combined objective prefers this shift when the weighted residual reduction exceeds the added denoising cost. If the favored solution lies outside the data support, this preference results in \emph{leakage}.}
\label{fig:main-figure-penalty}
\end{figure*}
Here $\lambda_t>0$ weights denoising and $c_t\ge0$ weights the residual, written as a constant scale $c\ge0$ times the fixed time-dependent schedule $\omega_t$ of \citet{bastek2025physics}. 
At fixed $t$, only their ratio $\eta_t$ matters and we write it as $\eta$ below. We use the clean sample $x_0$ parameterization \citep{karras2022elucidating}. 


\subsection{Setup}
We assume that the data samples satisfy the governing PDE, so
$R(x_0)=0$ almost surely for $x_0\sim\pdata$. Here $\pdata$ denotes the distribution the model is meant to learn. It is fixed by the application rather than by the equation, so it typically covers a restricted part of the solution set. We distinguish the support of the
data distribution $\M$ from the zero set of the residual $\Z$:
\[
  \M:=\supp(\pdata),\qquad \Z:=\{x\in\R^d:R(x)=0\}.
\]
We assume that $R$ is continuous, which gives $\M\subseteq\Z$. When the
inclusion is strict, $\Z\setminus\M$ contains fields that satisfy the PDE but
do not occur in the data distribution. 
While the governing PDE permits a broad family of solutions, the data of interest are collected under a restricted range of initial conditions, boundary conditions, and physical parameters, so that $\M\subsetneq\Z$ is the realistic setting.

\section{Analysis of the residual loss}
\label{sec:avoidance}

The residual vanishes at every feasible solution. However, the same prediction error can produce a large residual near one solution and a small residual near another. These differences matter when denoising is imperfect, as the combined objective may favor predictions near a solution that is less sensitive to errors, even if that solution is farther from the training target. We first derive the condition for this preference, then explain its relation to the residual Jacobian and the data distribution.

\subsection{Comparing predictions around two solutions}
\label{sec:ranking}

Fix a clean training sample $x_0\in\M$ and a noise level $t$. Dividing
the objective in Eq.~\eqref{eq:objective} by $\lambda_t$, the loss of a prediction $\hat x_0$ becomes
\begin{equation}
  \ell(\hat x_0) = \|\hat x_0 - x_0\|^2 + \eta\,\|R(\hat x_0)\|^2 .
  \label{eq:per_sample_loss}
\end{equation}
Here $\eta=c_t/\lambda_t$, with $\lambda_t>0$. To isolate the role of the residual term, we compare two predictions
that are centered at different feasible solutions but perturbed in the
same way. Let $y\in\Z$ be a second solution, so that
$R(y)=R(x_0)=0$. Let $\delta$ be a random perturbation
with $\E[\delta]=0$ and $\E\|\delta\|^2<\infty$.
We keep its distribution fixed and compare the predictions
$y+\delta$ and $x_0+\delta$, which differ only in their
centers. We assume that both expected squared residuals
are finite. For a given $\delta$, the prediction around $y$ attains
the lower loss if and only if
\begin{equation}
\begin{aligned}
  &\|y + \delta - x_0\|^2 + \eta\|R(y + \delta)\|^2 < {}\\
  &\qquad \|x_0 + \delta - x_0\|^2
    + \eta\|R(x_0 + \delta)\|^2 .
\end{aligned}
\label{eq:pointwise}
\end{equation}
We now compare the expected losses, requiring $\E_\delta[\ell(y+\delta)]< \E_\delta[\ell(x_0+\delta)]$. Using $\E[\delta]=0$, this is equivalent to
\begin{equation}
  \eta\,\E_\delta\!\left[
    \|R(x_0+\delta)\|^2-\|R(y+\delta)\|^2
  \right]
  >\|y-x_0\|^2.
  \label{eq:pairwise}
\end{equation}
Thus, predictions centered at $y$ have lower expected loss
precisely when the weighted reduction in expected squared
residual exceeds the additional denoising cost
$\|y-x_0\|^2$. Figure \ref{fig:main-figure-penalty} illustrates the comparison between denoising and residual losses. Appendix~\ref{app:pairwise-derivation} provides a detailed derivation.

\subsection{Why the residual cost depends on the solution}
\label{sec:sensitivity}

To understand why the expected residual loss can differ
between the two solutions, consider a small perturbation
around a feasible solution. Assume that $R$ is twice continuously differentiable near $x_0$ and $y$. Since $R(y)=0$, Taylor expansion gives
\[
  R(y+\delta)=J_R(y)\delta+O(\|\delta\|^2).
\] 
Each entry of the \emph{residual Jacobian} $J_R(y)$ measures how one residual component changes when one component of the field is varied, so $J_R(y)\,\delta$ is the first-order response of the residual to the perturbation $\delta$. The Jacobian can differ between solutions, so the same perturbation can produce different residual responses
around $x_0$ and $y$.

To control the approximation after averaging, consider $\delta=\epsilon z$, where $z$ is bounded, and $\epsilon$ is small. Under the smoothness assumption above, replacing the residual by its linear approximation changes its expected squared norm by $O(\epsilon^3)$. Applying this approximation to both solutions gives the approximate comparison
\[
  \eta\,\E_\delta\!\left[
    \|J_R(x_0)\delta\|^2-\|J_R(y)\delta\|^2
  \right]
  >\|y-x_0\|^2.
\]
The difference in expected residual loss between the two centers depends on how strongly their Jacobians respond to the same perturbations. A smaller average response at $y$ reduces the expected residual penalty. The approximate comparison
predicts a lower combined loss only when this reduction, weighted by $\eta$, exceeds the displacement cost $\|y-x_0\|^2$. 
We call this preference for less sensitive solutions \emph{avoidance}. 

\subsection{Feasibility does not imply data support}
\label{sec:leakage}

The comparison in Sec.~\ref{sec:ranking} requires only that the competing solution is feasible, $R(y)=0$, not that it is supported by the data. Every element of $\Z$ meets this requirement, so the competing solution can also lie in $\Z\setminus\M$. We refer to the possibility that the combined objective favors such a solution as \emph{leakage}. 

The next section addresses avoidance and leakage by fixing the residual Jacobian at the clean training sample, so that moving the prediction cannot change the Jacobian used to
penalize its error.

\section{Data-anchored residual loss}
\label{sec:method}

The preceding analysis shows that evaluating the residual at the prediction alone can favor different solutions with lower residual sensitivity, including solutions outside the data support. The penalty can therefore decrease as the prediction moves away from its clean training target. To remove this incentive, we penalize the prediction error using the residual Jacobian evaluated at the clean sample $x_0$. 

Since $R(x_0)=0$, the first-order Taylor approximation of the residual
at a prediction $\hat x_0$ is
\[
  R(\hat x_0)\approx J_R(x_0)(\hat x_0-x_0).
\]
We use the squared norm of this approximation as the
\emph{data-anchored residual loss}:
\begin{equation}
  \|J_R(x_0)(\hat x_0-x_0)\|^2.
  \label{eq:anchored-plain}
\end{equation}
For a given training sample, the Jacobian remains fixed as the prediction changes. The prediction therefore cannot obtain a smaller penalty simply by moving toward a solution with a less sensitive Jacobian.

\paragraph{Normalization.}
\label{sec:normalization}

The magnitude of the residual Jacobian can vary across training samples. This can make the anchored penalty much larger for some samples than others, even when their prediction errors have the same size. To control this scale, we divide by the squared Frobenius norm of the Jacobian and write the full training objective as:
\begin{equation}
\begin{aligned}
  \mathcal L_{\mathrm{anch}}
  =\E_{t,x_0,x_t}\!\left[
    \lambda_t\|\delta\|^2+
    c_t\frac{\|J_R(x_0)\delta\|^2}
             {\|J_R(x_0)\|_F^2+\gamma}
  \right],\\
  \delta=\hat x_0-x_0,\qquad
  c_t=c\,\lambda_t,\qquad \gamma>0.
\end{aligned}
\label{eq:anchored-objective}
\end{equation}
The constant $\gamma$ prevents division by zero when the Jacobian vanishes. We use the denoising weight as the schedule of the residual term, $c_t=c\,\lambda_t$, so the ratio $\eta_t=c_t/\lambda_t=c$ is the same at every noise level.

This normalization gives a direct bound:
\[
  0\leq
  \frac{\|J_R(x_0)\delta\|^2}
       {\|J_R(x_0)\|_F^2+\gamma}
  \leq \|\delta\|^2.
\]
Thus the normalized residual term cannot exceed the squared prediction error before the loss weights are applied.

\begin{figure*}[t!]
\centering
\includegraphics[width=0.94\textwidth]{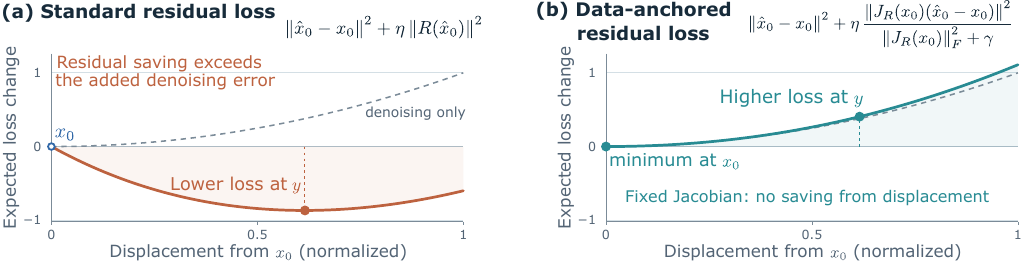}
\caption{Comparison of the standard and data-anchored objectives for the displacement illustrated in Figure~\ref{fig:main-figure-penalty}. Curves show the change in expected loss relative to the predictions at $x_0$, with the perturbation distribution held fixed. The dashed curve shows the added denoising cost. (a) The standard residual penalty can decrease enough to outweigh this cost, favoring predictions centered at $y$. (b) Anchoring the residual Jacobian at $x_0$ removes this saving, so displacement can only increase the expected combined loss, whose minimum remains at $x_0$.}
\label{fig:main-figure-losses}
\end{figure*}

\subsection{Effect on avoidance}
\label{sec:anchoring}

We now repeat the comparison of Section~\ref{sec:ranking} for the anchored loss in Equation~\ref{eq:anchored-objective} and show that a prediction centered away from $x_0$ can no longer attain a lower expected loss. As before, we compare the predictions $y+\delta$ and $x_0+\delta$, which share the same perturbation $\delta$ with $\E[\delta]=0$. The prediction around $y$ therefore attains the lower expected loss if and only if
\begin{equation} -\eta\,\frac{\|J_R(x_0)(y-x_0)\|^2}{\|J_R(x_0)\|_F^2+\gamma}>\|y-x_0\|^2, \label{eq:anchored-relocation} \end{equation}
which is the anchored counterpart of Equation~\ref{eq:pairwise}. The left-hand side is never positive, since it is the negative of a squared norm divided by a positive constant, while the right-hand side is strictly positive for $y\neq x_0$. The condition therefore never holds, and $x_0$ is the unique minimizer of the expected anchored loss over all centers. Appendix~\ref{app:anchored-relocation} provides a detailed derivation.

Intuitively, the standard loss measures the sensitivity of the residual at the prediction, whereas the anchored loss always measures it at the clean training sample. For a prediction around $y$, the standard penalty is therefore governed by $J_R(y)$, so a less sensitive solution can reduce it enough to outweigh the added denoising cost (Section~\ref{sec:sensitivity}). The anchored penalty uses $J_R(x_0)$ for both predictions, so the shared perturbation incurs the same expected cost around either center, and moving the center can only add to the penalty. Figure~\ref{fig:main-figure-losses} illustrates this behavior and compares the anchored loss with the standard loss.

\subsection{Alignment with the denoising gradient}
\label{sec:gradient-alignment}

Reducing the PDE residual does not necessarily bring a prediction closer to its clean target. Anchoring prevents this conflict at the level of prediction-space gradients. For a fixed feasible target $x_0$, the gradients of the standard and normalized anchored physics penalties
with respect to $\hat x_0$ are
\begin{equation}
\begin{aligned}
  g_{\mathrm{std}}
  &=2J_R(\hat x_0)^\top R(\hat x_0),\\
  g_{\mathrm{anch}}
  &=\frac{2J_R(x_0)^\top J_R(x_0)(\hat x_0-x_0)}
          {\|J_R(x_0)\|_F^2+\gamma}.
\end{aligned}
\label{eq:physics-gradients}
\end{equation}
The standard gradient depends on the residual geometry at the prediction. The anchored gradient instead weights the prediction error using the fixed local sensitivity at the clean target.

Taking its inner product with the denoising gradient
$g_{\mathrm{den}}=2(\hat x_0-x_0)$ gives
\begin{equation}
  \langle g_{\mathrm{den}},g_{\mathrm{anch}}\rangle
  =
  \frac{4\|J_R(x_0)(\hat x_0-x_0)\|^2}
       {\|J_R(x_0)\|_F^2+\gamma}
  \geq0.
  \label{eq:anchored-gradient-alignment}
\end{equation}
Thus an infinitesimal descent step on the anchored penalty cannot increase the squared prediction error. The standard residual gradient has no such guarantee. 

Anchoring also preserves the leading local residual gradient. Since $R(x_0)=0$, Taylor expansion of $R$ and its Jacobian around $x_0$ gives
\begin{equation}
\begin{aligned}
\hspace{-1.5mm}  g_{\mathrm{std}}
  &=
  \bigl(\|J_R(x_0)\|_F^2+\gamma\bigr)
  g_{\mathrm{anch}} + O(\|\hat x_0-x_0\|^2).
\end{aligned}
\label{eq:local-gradient-agreement}
\end{equation}
The anchored gradient, therefore, agrees with the standard residual gradient to first order, up to normalization, while remaining aligned with denoising even away from the target.

\section{Experiments}

\paragraph{Datasets.}
We use one synthetic dataset, three PDE systems, and one molecular dataset.

The \emph{shocked circle} is a synthetic dataset whose samples are uniformly distributed around a unit circle. We embed this circle in $\R^D$ by placing it in the first two coordinates and setting all remaining coordinates to zero (App.~\ref{app:circle}). We define the residual as $R(x)=(\|x\|^2-1)\,g(x)$, where $g$ is positive. The residual therefore vanishes at every point on the unit sphere $\mathbb S^{D-1}$. The factor $g$ controls how sensitive the residual is to prediction errors without changing this feasible set. Along the data circle, the sensitivity is $\sigma=2g/\sqrt D$. We use a \texttt{graded} profile, which makes sensitivity vary by a factor of $101$ around the circle, and a \texttt{flat} control with $g\equiv1$ and constant sensitivity. Comparing these profiles isolates the effect of varying residual sensitivity. We use $D=10$ to examine the role of feasible states outside the data support. The data lie on the unit circle in the plane of the first two coordinates, while the residual vanishes on the whole hypersphere. This allows us to study leakage toward feasible states absent from the data.

\emph{Advection} is the linear 1D advection equation $\partial_\tau u+\beta\partial_\xi u=0$ with $\beta=1$ on a $256\times256$ space--time grid. Its residual is linear in $u$, so the residual Jacobian is the same for every sample. 

\emph{Burgers} is the nonlinear 1D viscous Burgers equation $\partial_\tau u+u\,\partial_\xi u-\nu\partial_{\xi\xi}u=0$ with $\nu=10^{-3}$ on a $256\times512$ space--time grid. Its trajectories develop steep fronts (shocks), and its residual Jacobian depends on the state through the nonlinear term $u\,\partial_\xi u$. The two time-dependent PDE datasets, advection and Burgers, are generated with \texttt{torchfsm} \citep{liu_2025_15350210} on the periodic domain $\xi\in[0,1)$ from random multi-mode initial conditions. Each contains $10\,000$ trajectories, split $90/10$ into training and held-out data, with each sample representing a full space--time trajectory (App.~\ref{app:datasets}).

\emph{Darcy} describes steady flow through a porous medium, $-\nabla\cdot(K\nabla p)=f$ on the unit square, with permeability $K$, pressure $p$, zero-flux boundaries, and a source and a sink in two opposite corners. The log-permeability is a Gaussian random field with length scale $0.1$, and $p$ is the solution for that $K$. The model generates $K$ and $p$ jointly on a $64\times64$ grid.

For the molecular task, we evaluate the generation of transition states (TSs) of chemical reactions on \emph{Transition1x-2p3p4p} \citep{darouich2026robust}. The dataset contains about $14\,000$ gas-phase reactions of small organic molecules with up to 7 heavy atoms and up to 23 atoms, including elements up to the fourth period. All structures are computed at the semi-empirical GFN2-xTB level of theory \citep{bannwarth2019gfn2}.

\begin{figure*}[!t]
\centering
\includegraphics[width=0.98\textwidth]{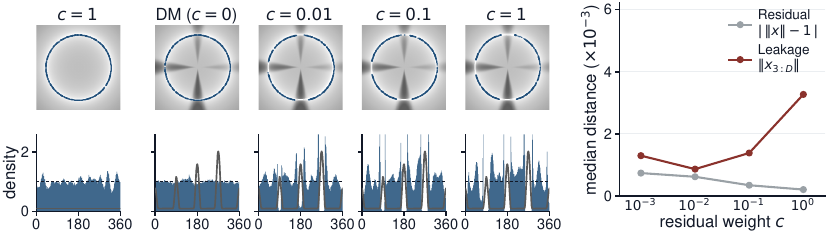}
\caption{PIDM on the shocked circle at $D=10$. Top, generated samples projected onto the data plane, with shading indicating residual sensitivity. Bottom, their angular density compared with the data distribution (dashed) and the residual sensitivity $\sigma(\theta)$ (grey). The first column uses the \texttt{flat} residual as a control. Right, median radial residual $|\|x\|-1|$ and distance to the data plane $\|x_{3:D}\|$ as functions of the penalty weight.}
\label{fig:toy}
\end{figure*}

\paragraph{Metrics.}
On the shocked circle, we report the median distance to the feasible sphere, $|\|x\|-1|$, and to the data plane, $\|x_{3:D}\|$. The latter measures departures from the plane containing the data and serves as a diagnostic for leakage.

On the PDE datasets, physical accuracy is measured by the mean absolute residual $\overline{|R|}$, averaged over residual entries and generated samples. Maximum mean discrepancy (MMD) assesses distributional fidelity by comparing the joint distributions of field features, including amplitude and maximum gradient, between generated and held-out samples. The $\sigma$ ratio compares their median residual sensitivities, with values below one indicating less sensitive generated states. The field standard-deviation ratio compares their mean within-field standard deviations, using $u$ for Burgers and advection and $p$ for Darcy. Both ratios ideally equal one.

For molecular TSs, we compare each generated structure with the reference TS of the same reaction. RMSD and DMAE measure structural errors. Physical validity is assessed by the energy difference $\Delta E$, the maximum atomic force $F_{\max}$, the Hessian MAE, and the valid TS rate, defined as the percentage of generated structures whose Hessian has exactly one negative eigenvalue. Appendix~\ref{app:metrics} gives the full definitions.

\paragraph{Weights.}
The numerical penalty weights are not directly comparable between PIDM and the anchored objective. PIDM penalizes the residual in its original scale, whereas the anchored objective divides its linearized residual penalty by the squared Frobenius norm of the residual Jacobian. A larger anchored coefficient can compensate for this normalization rather than impose a stronger constraint. To compare the methods, we approximately match their local penalty scales near the training data, accounting for both Jacobian normalization and different weighting schedules across noise levels. This gives the pairings $c_{\mathrm A}\approx k c_{\mathrm P}$, with conversion factors of $2\times10^6$ for advection, approximately $10^8$ for Burgers, and approximately $10^{10}$ for Darcy. We evaluate a range of weights to examine the resulting trade-offs between physical accuracy and data fidelity.

\begin{figure*}[t]
    \centering
    \includegraphics[width=\textwidth]{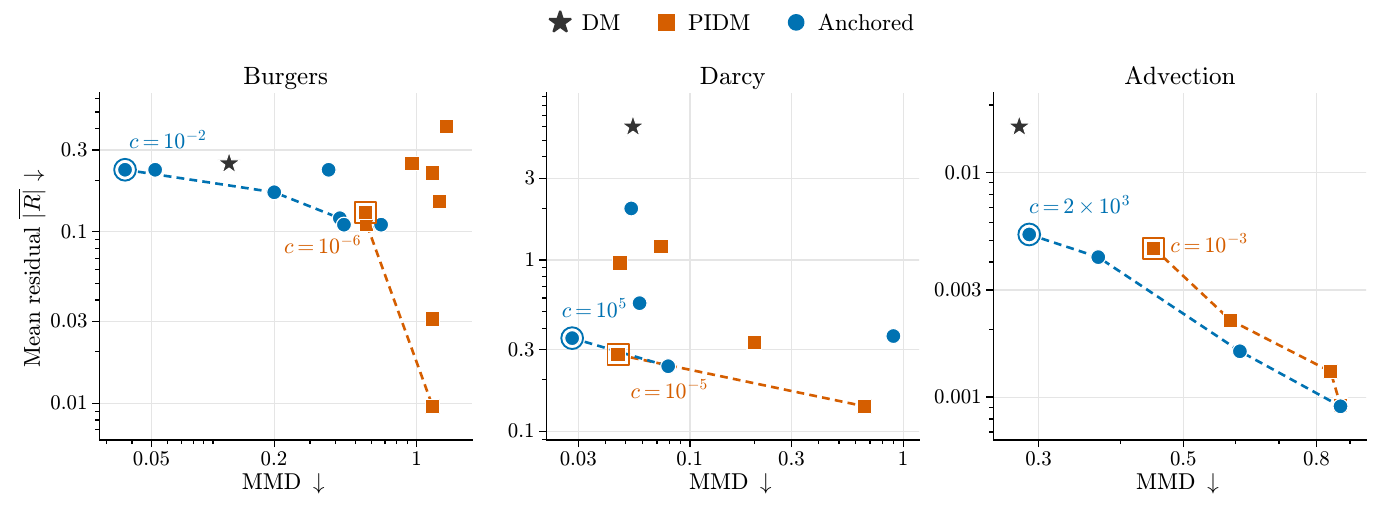}
    \caption{
        Trade-offs between distributional fidelity and physical accuracy on Burgers (left), Darcy (middle), and advection (right). Each point shows the MMD and mean absolute residual
        $\overline{|R|}$ for one reported setting. Both axes are logarithmic, and lower left is better. Stars denote DM, orange squares PIDM, and blue circles the anchored objective. Dashed lines connect nondominated settings within each regularized method. A setting is nondominated if no other tested weight of the same method reduces one metric without increasing the other. The line segments serve as visual guides. Rings and labels mark the lowest-MMD setting for each regularized
        method, breaking MMD ties by lower residual. The labeled $c$ values are the coefficients used in each method's objective. 
    }
    \label{fig:pde-pareto}
\end{figure*}

\subsection{Results}

\paragraph{Shocked circle.} 
\label{sec:exp-toy}

Figure~\ref{fig:toy} shows PIDM on the \texttt{graded} circle at
$D=10$. Darker shading indicates higher residual sensitivity. Errors of the same size incur larger residuals. As $c$ increases,
the residual decreases, but samples increasingly avoid sensitive
regions. The model thus favors states where the same error is
penalized less, as predicted by Eq.~\eqref{eq:pairwise}.
This effect is absent in the \texttt{flat} control, where sensitivity
is constant and the angular density remains uniform even at $c=1$.
Meanwhile, the distance to the data plane increases despite the
lower residual. Since all data lie in this plane, this distance
measures departure from the data manifold. PIDM therefore lowers
the residual both by avoiding sensitive regions and by moving
along $\Z\setminus\M$, where the residual vanishes and cannot
distinguish data-supported states from other feasible states.
The anchored objective retains coverage of the entire circle
(App.~\ref{app:toy_detail}).

\paragraph{Burgers' equation.}
\label{sec:exp-burgers}
Only the anchored objective dominates DM. At $c=10^{-2}$ it lowers the residual from $0.25$ to $0.23$ and the MMD from $0.12$ to $0.037$, and its $\sigma$ and $\mathrm{std}(u)$ ratios rise from $0.82$ and $0.81$ to $0.91$ and $0.90$. It thus counteracts the tendency of DM toward smooth fields. No weight of PIDM reaches the MMD of DM. At its smallest weights PIDM lowers the residual to $0.11$ at an MMD of $0.57$, and larger weights shrink the fields until the $\mathrm{std}(u)$ ratio falls below $0.1$. Its very low residuals at $c\geq10^{-1}$ therefore come from weak, smooth fields with an MMD of $1.2$. The anchored objective reaches the same residual of $0.11$ at an MMD of $0.44$ and keeps an $\mathrm{std}(u)$ ratio of at least $0.63$ up to $c=10^{4}$. Wherever both methods reach the same residual, the anchored samples are closer to the data. Detailed results can be found in Appendix~\ref{app:burgers_detail}.

\paragraph{Darcy's equation.}
\label{sec:exp-darcy}

Both objectives match the MMD of DM and lower its residual from $6.0$ to $0.28$ for PIDM and $0.35$ for the anchored objective. The anchored objective reaches the lowest MMD of all runs ($0.028$ against $0.046$ for PIDM and $0.054$ for DM), with $\sigma$ and $\mathrm{std}(p)$ ratios closer to one. At larger weights PIDM flattens the pressure, which makes errors in $K$ cheaper, and its MMD rises to $0.20$ at $c=10^{-4}$ (Table~\ref{tab:darcy}). At the comparable weight $c=10^{6}$ the anchored objective still keeps the pressure, with a $\mathrm{std}(p)$ ratio of $0.92$, and reaches a residual of $0.24$ at an MMD of $0.079$. At the largest weights both objectives flatten the pressure (App.~\ref{app:darcy_detail}).

\begin{table}[t!]
\vspace{-2.5mm}
\centering
\small
\setlength{\tabcolsep}{3pt}
\caption{Results at the weight $c$ that minimizes the residual subject to MMD being no higher than the DM baseline. ``--'' indicates that no tested weight can match the MMD of the DM baseline. Bold marks the best-displayed value within each PDE, with ratios compared by their distance from it. The field standard-deviation ratio uses $u$ for Burgers and $p$ for Darcy.}
\label{tab:pde-selected}
\begin{tabular*}{\columnwidth}{@{\extracolsep{\fill}}lrrrrr@{}}
\toprule
Method & $c$
& \shortstack{$\overline{|R|}$\\$\downarrow$}
& \shortstack{MMD\\$\downarrow$}
& \shortstack{$\sigma$ ratio\\$\to 1$}
& \shortstack{Field std.\\ratio $\to 1$} \\
\midrule
\multicolumn{6}{@{}l}{\textit{Burgers}} \\
\addlinespace[2pt]
DM       & --        & $0.25$          & $0.12$           & $0.82$          & $0.81$          \\
PIDM     & None      & --              & --               & --              & --              \\
Anchored (ours) & $10^{-2}$ & $\mathbf{0.23}$ & $\mathbf{0.037}$ & $\mathbf{0.91}$ & $\mathbf{0.90}$ \\
\midrule
\multicolumn{6}{@{}l}{\textit{Darcy}} \\
\addlinespace[2pt]
DM       & --        & $6.0$           & $0.054$          & $0.84$          & $1.1$           \\
PIDM     & $10^{-5}$ & $\mathbf{0.28}$ & $0.046$          & $0.83$          & $0.98$          \\
Anchored (ours) & $10^{5}$  & $0.35$          & $\mathbf{0.028}$ & $\mathbf{0.90}$ & $\mathbf{0.99}$ \\
\bottomrule
\end{tabular*}
\end{table}

\paragraph{Advection equation.}
\label{sec:exp-advection}

For the linear advection equation the residual Jacobian is the same for every field, so the
anchored residual equals the standard one up to a constant, and neither avoidance nor a
shifted target can occur (App.~\ref{app:anchored}). The two objectives then differ only in
their weighting over noise levels. As expected, they behave alike
(Figure~\ref{fig:pde-pareto}). Both lower the residual at the cost of amplitude, and at the
largest weights they coincide (Table~\ref{tab:advection}). The smaller loss of amplitude of the anchored objective at
low weights comes from PIDM's ratio $\eta_t$ growing at high noise, where the zero field is
an exact solution.

\begin{table*}[t]
\centering\small
\setlength{\tabcolsep}{3pt}
\caption{TS generation on Transition1x-2p3p4p. All metrics except the valid TS rate are medians over the generated structures. The Hessian MAE is computed against the GFN2-xTB reference Hessians, and the valid TS rate is the percentage of generated structures whose Hessian has exactly one negative eigenvalue.}
\label{tab:structural_metrics_summary}
\begin{tabular}{@{}lrrrrrr@{}}
\toprule
 & \shortstack{RMSD\\(\AA)$\downarrow$} & \shortstack{D-MAE\\(\AA)$\downarrow$} & \shortstack{$\Delta E$ \\(kcal/mol)$\downarrow$} & \shortstack{$F_\text{max}$ \\(kcal/mol/\AA)$\downarrow$} & \shortstack{Hessian MAE\\(kcal/mol/\AA$^2$)$\downarrow$} & \shortstack{Valid TS\\(\%)$\uparrow$} \\
\midrule
FM & \textbf{0.134} & 0.060 & 2.53 & 33.50 & 8.40 & 57.0 \\
FM + residual & 0.140 & 0.066 & 8.11 & 67.80 & 12.56 & 36.0 \\
FM + anchored (ours) & 0.135 & \textbf{0.059} & \textbf{2.36} & \textbf{31.22} & \textbf{8.09} & \textbf{60.0} \\
\bottomrule
\end{tabular}
\end{table*}

\paragraph{Transition1x-2p3p4p.}
\label{sec:exp-ts}

To test the anchored loss beyond PDEs, we generate TSs of chemical reactions. A TS is a first-order saddle point of the PES, at which all forces vanish, so a natural residual is the force $R(\hat x_0)=F(\hat x_0)=-\nabla E(\hat x_0)$. Since evaluating GFN2-xTB during training is too expensive, we use the machine-learned interatomic potential HIP \citep{burger2025shoot} as a surrogate, which provides energies, forces, and Hessians at a fraction of the cost.

We train React-OT \citep{duan2025optimal} with the plain flow matching loss (FM), with the standard residual loss added to it (FM + residual), and with the anchored residual loss added to it (FM + anchored), as detailed in Appendix~\ref{app:implementation}. All three models reach similar structural accuracy (Table~\ref{tab:structural_metrics_summary}). RMSD and DMAE weigh all geometric errors equally, but the energy is far more sensitive to some displacements, such as bond stretches, than to others, such as torsions. As with shocked and smooth PDE solutions, errors of the same size can therefore have very different physical costs, and the energetic metrics separate the models clearly. Although FM + residual penalizes the force directly, it roughly triples the energy error of FM, doubles its maximum force, and increases its Hessian error, while FM + anchored improves all three. The valid TS rate drops from $57\%$ for FM to $36\%$ for FM + residual and rises to $60\%$ for FM + anchored. This is consistent with leakage, since the standard residual cannot distinguish a TS from a minimum, whereas the anchored loss evaluates the Jacobian at the reference TS, whose Hessian carries the single direction of negative curvature.

\section{Related work}
\label{sec:related}

\paragraph{Physics at inference.} One line of work integrates physics
during inference and steers the samples of a trained model with the
gradients of the PDE residual
\citep{shu2023physics,huang2024diffusionpde,jacobsen2025cocogen,gallon2026physicsinformed}, by projections onto the constraint set \citep{cheng2025gradient,utkarsh2025physics,xu2026right}, or by running the diffusion in a transformed space in which the map back to the data space carries the constraint \citep{liu2023mirror,feng2025neural}. For diffusion guidance in general, \citet{geyfman2026calibrated} show that common approximations can bias sampling away from the intended Bayesian posterior. We instead study residual penalties during training, with sampling left unchanged.

\paragraph{Physics during training.} \citet{bastek2025physics} introduce the
penalty we study, adding the residual loss to the
denoising loss. Later work varies how the residual enters. \citet{zeng2025physicsinformed}
propose a different weighting of the residual across noise levels,
\citet{sawhney2025beyond} pass the residual field to spectral attention instead of
the loss, and \citet{yeom2025softly} add the constraint gradient to the
predicted clean sample inside the forward pass.
Another group penalizes the residual on a sampled output
rather than on the posterior mean, by distilling the trained model into a few-step
generator or fine-tuning it on multi-step trajectories
\citep{zhang2026physicsinformed,kong2026ultra,chang2026stabilizing}, or by treating
the residual as a reward and fine-tuning a pretrained model against it
\citep{yuan2025pirf,tauberschmidt2026physics}, at the cost of a second training stage. These corrections reduce the prediction error but leave the residual evaluated at the model's own output, so the preference derived (Sec.~\ref{sec:ranking}) persists whenever the prediction is not exact. 
An extended overview over related work can be found in App.~\ref{app:related_work}.

\section{Discussion}
\paragraph{Limitations.} The anchored objective requires a residual Jacobian--vector product and estimates the normalizer at every training step, potentially increasing the computational cost relative to PIDM (Table~\ref{tab:cost}). Its benefit depends on the residual structure. For linear PDEs such as advection, the anchored and standard penalties coincide up to a constant, leaving only their weighting across noise levels to distinguish the objectives. The gains are clearest when the residual depends nonlinearly on the state. Anchoring also guarantees neither exact constraint satisfaction nor preservation of the data distribution. It mitigates avoidance, but sample-dependent Jacobian weighting can still shift the conditional denoising optimum (App.~\ref{app:anchored}). Empirically, generated samples retain residuals above those of the data, and large penalty weights still smooth Burgers shocks and flatten Darcy pressure fields, although anchoring reduces these distortions over much of the tested range of weights. The penalty weight, therefore, remains an important choice in balancing physical accuracy and distributional fidelity.

\paragraph{Future work.} Since the anchored objective keeps the samples close to the data distribution, it is a good starting point for a correction after sampling, such as a final projection, which could provide the guarantees that training alone cannot. A time weighting chosen for the physics term, rather than inherited from the data term, is also likely to improve the method. Further directions are efficient approximations of the Jacobian and applying the physics term only at the noise levels where it helps most.

\subsection*{AI use statement}
In this work, we used generative AI tools 
to aid and polish writing, 
to draft sections of the paper, 
and to support research execution through code generation for analysis and evaluation scripts, generation of datasets and creation of figure.
We have not used generative AI tools for 
design methodology and experiments, 
develop theoretical models or conceptual frameworks 
and assist with translation, 
clean and reformat dataset are not applicable to this work, summarize and analyze existing literature,  format references.
Additionally, we used generative AI tools for creating and editing figures and draft part of the research paper, search for additional relevant literature.
We have reviewed all AI-assisted work. All code and text has been manually reviewed by the authors.
We take responsibility for the final content of this work, including text, claims, or artifacts produced with the aid of generative AI.

See the AISTATS 2027 AI Policy for Authors for more details. This statement should not be more than 1 page.

\subsubsection*{Acknowledgements}
This work was supported by the Deutsche Forschungsgemeinschaft (DFG, German Research Foundation) under Germany’s Excellence Strategy – EXC 2120/1 – 390831618.
The authors thank the International Max Planck Research School for Intelligent Systems (IMPRS-IS) for supporting Tanja Bien, Samir Darouich, Lluís Pastor-Pérez, and Vinh Tong.
We also acknowledge support from the Ministry of Science, Research and the Arts Baden-Württemberg through the Artificial Intelligence Software Academy (AISA), as well as from the Stuttgart Center for Simulation Science (SimTech).
L.\ Mualem was supported by a postdoctoral scholarship from the Planning and Budgeting Committee (PBC) of the Council for Higher Education in Israel. 
The authors gratefully acknowledge the computing time provided on the high-performance computer HoreKa by the National High-Performance Computing Center at KIT (NHR@KIT). This center is jointly supported by the Federal Ministry of Education and Research and the Ministry of Science, Research and the Arts of Baden-Württemberg, as part of the National High-Performance Computing (NHR) joint funding program (https://www.nhr-verein.de/en/our-partners). HoreKa is partly funded by the German Research Foundation (DFG). 

\bibliography{bibliography}

\section*{Checklist}
\begin{enumerate}

  \item For all models and algorithms presented, check if you include:
  \begin{enumerate}
    \item A clear description of the mathematical setting, assumptions, algorithm, and/or model. Yes, all assumptions and definitions are stated in Section~\ref{sec:avoidance} and \ref{sec:method} and Appendix~\ref{app:displacement}, \ref{app:pairwise-derivation}, \ref{app:anchored-relocation} include the detailed step-by-step derivations and more details. 
    \item An analysis of the properties and complexity (time, space, sample size) of any algorithm. Yes, Appendix~\ref{app:cost} analysis the training cost of the different methods. 
    \item (Optional) Anonymized source code, with specification of all dependencies, including external libraries. Yes.
  \end{enumerate}

  \item For any theoretical claim, check if you include:
  \begin{enumerate}
    \item Statements of the full set of assumptions of all theoretical results. Yes
    \item Complete proofs of all theoretical results. Yes
    \item Clear explanations of any assumptions. Yes
  \end{enumerate}

  \item For all figures and tables that present empirical results, check if you include:
  \begin{enumerate}
    \item The code, data, and instructions needed to reproduce the main experimental results (either in the supplemental material or as a URL). Yes, code and data are added as supplemental material.
    \item All the training details (e.g., data splits, hyperparameters, how they were chosen). Yes, in Appendix~\ref{app:implementation} Implementation details.
    \item A clear definition of the specific measure or statistics and error bars (e.g., with respect to the random seed after running experiments multiple times). Yes, in Appendix~\ref{app:metrics} Metrics
    \item A description of the computing infrastructure used. (e.g., type of GPUs, internal cluster, or cloud provider). Yes, we name the type of GPU used for the respective experiments in the training and sampling settings table.
  \end{enumerate}

  \item If you are using existing assets (e.g., code, data, models) or curating/releasing new assets, check if you include:
  \begin{enumerate}
    \item Citations of the creator If your work uses existing assets. Yes, we cite all datasets and software we use.
    \item The license information of the assets, if applicable. Yes
    \item New assets either in the supplemental material or as a URL, if applicable. Not Applicable.
    \item Information about consent from data providers/curators. Not Applicable.
    \item Discussion of sensible content if applicable, e.g., personally identifiable information or offensive content. Not applicable
  \end{enumerate}

  \item If you used crowdsourcing or conducted research with human subjects, check if you include:
  \begin{enumerate}
    \item The full text of instructions given to participants and screenshots. Not Applicable
    \item Descriptions of potential participant risks, with links to Institutional Review Board (IRB) approvals if applicable. Not Applicable
    \item The estimated hourly wage paid to participants and the total amount spent on participant compensation. Not Applicable
  \end{enumerate}

\end{enumerate}

\clearpage
\appendix
\thispagestyle{empty}

\onecolumn
\aistatstitle{Appendix}

\begin{center}
    \startcontents[sections]\vbox{\vspace{4mm}\sc
    \printcontents[sections]{l}{0}{\setcounter{tocdepth}{2}}}
\end{center}

\newpage

\section{Conditional target of the combined loss} 
\label{app:displacement}
In physics-constrained training, the governing equation satisfied by the data
is known, and its residual is added to steer predictions toward physically
plausible samples \cite{bastek2025physics}. Consider the idealized case in which the
diffusion model already reproduces $\pdata$. Its generated samples then
satisfy the PDE, so the residual has no distributional defect to correct.
We ask whether adding the residual loss nevertheless changes the optimal denoiser's predictions. 

To answer this, we compare the minimizer of the denoising loss with the minimizer of the combined loss. At a fixed noisy input $x_t$, the first is the conditional mean $x_0^\star(x_t)=\E[x_0\mid x_t]$, and the second, is
\begin{equation}
  x_\eta^\star(x_t)\in\argmin_{y\in\R^d}
  \Bigl\{\underbrace{\tfrac12\|y-x_0^\star\|^2}_{\text{denoising loss}}
    +\underbrace{\tfrac\eta2\|R(y)\|^2}_{\text{residual loss}}\Bigr\}.
  \label{eq:population-target}
\end{equation}
The displacement from the denoising optimum to the combined optimum is $s_\eta(x_t)\coloneqq x_\eta^\star(x_t)-x_0^\star$, with $s_0(x_t)=0$ when only the denoising objective is used. 

To find the minimizer of Eq.~\ref{eq:population-target} we ask how far a prediction must shift from $x_0^\star$ to reach it, so we write every candidate as $y=x_0^\star+s$ and search for the displacement $s$ at which the gradient of the combined objective vanishes. 
Setting the gradient of Eq.~\ref{eq:population-target} to zero gives a condition for $s_\eta$, and its Taylor expansion around $x_0^\star$ to leading order in $\eta$ is (App.~\ref{app:displacement-condition}, App.~\ref{app:small-weight})
\begin{equation}
  s_\eta=-\eta\,J_R(x_0^\star)^{\!\top}R(x_0^\star)+O(\eta^2)
        =-\tfrac\eta2\nabla\|R\|^2\big|_{x_0^\star}+O(\eta^2).
  \label{eq:population-shift}
\end{equation}
If the product of the Jacobian of the residual and the residual is nonzero $J_R(x_0^\star)^{\!\top}R(x_0^\star)\neq0$, the denoising
optimum cannot minimize the combined objective for any $\eta>0$.
Thus $s_\eta\neq0$, even though $x_0^\star$ is already the optimal
denoising prediction. This can occur when every clean sample satisfies
the PDE, because their conditional mean need not satisfy a nonlinear
equation.

The remainder of this section derives these statements. App.~\ref{app:conditional-loss} reduces the conditional denoising objective to
Eq.~\ref{eq:population-target}, App.~\ref{app:displacement-condition} and
App.~\ref{app:small-weight} give the stationarity condition and its expansion for
small weight, App.~\ref{app:displacement-bound} bounds the magnitude of $s_\eta$,
and App.~\ref{app:conditional-gap} examines when the conditional mean of feasible
samples fails to satisfy the equation. App.~\ref{app:latitude} gives an example in
which every increase of the weight lowers the residual and moves the target
farther from the data. App.~\ref{sec:jensen} relates the infeasibility of the
conditional mean to the Jensen gap.

\subsection{The conditional denoising objective}
\label{app:conditional-loss}

Fix $x_t$ and write $x_0^\star=x_0^\star(x_t)$ and
$\Sigma_\post(x_t)=\Cov[x_0\mid x_t]$. With finite conditional second
moments, expand $x_0-y=(x_0-x_0^\star)+(x_0^\star-y)$. Since
$\E[x_0-x_0^\star\mid x_t]=0$, the cross term has zero expectation and
\begin{equation}
  \E[\|x_0-y\|^2\mid x_t]
  =\|y-x_0^\star\|^2+\tr\Sigma_\post(x_t).
\end{equation}
The covariance term is independent of $y$, giving the objective in
Eq.~\ref{eq:population-target}.

\subsection{The displacement condition}
\label{app:displacement-condition}

Let $R=(R_1,\ldots,R_m)$ be
differentiable. Its Jacobian $J_R(y)\in\R^{m\times d}$ has entries
$[J_R(y)]_{ki}=\partial R_k(y)/\partial y_i$. Write a candidate prediction as
$y=x_0^\star+s$. After this substitution, the function minimized inside the
braces in Eq.~\ref{eq:population-target} becomes
\begin{equation}
  f_\eta(s)=\tfrac12\|s\|^2
    +\tfrac\eta2\|R(x_0^\star+s)\|^2.
\end{equation}
The derivative of the denoising term is
\begin{equation}
  \nabla_s\tfrac12\|s\|^2=s.
\end{equation}
For the residual term, write its squared norm componentwise as
\begin{equation}
  \tfrac12\|R(x_0^\star+s)\|^2
  =\tfrac12\sum_{k=1}^m R_k(x_0^\star+s)^2.
\end{equation}
Because $y=x_0^\star+s$, differentiation with respect to $s_i$ gives
\begin{equation}
  \frac{\partial}{\partial s_i}
  \tfrac12\|R(x_0^\star+s)\|^2
  =\sum_{k=1}^m R_k(x_0^\star+s)
    \frac{\partial R_k}{\partial y_i}(x_0^\star+s).
\end{equation}
Collecting these components yields
\begin{equation}
  \nabla_s\tfrac12\|R(x_0^\star+s)\|^2
  =J_R(x_0^\star+s)^{\!\top}R(x_0^\star+s).
\end{equation}
The gradient of the full objective is therefore
\begin{equation}
  \nabla_s f_\eta(s)
  =s+\eta J_R(x_0^\star+s)^{\!\top}R(x_0^\star+s).
\end{equation}
At the minimizing displacement $s=s_\eta$, the first-order condition is
\begin{equation}
  0=s_\eta+ \eta J_R(x_0^\star+s_\eta)^{\!\top}
       R(x_0^\star+s_\eta).
\end{equation}
Rearranging gives the implicit stationarity condition
\begin{equation}
  s_\eta=-\eta J_R(x_0^\star+s_\eta)^{\!\top}R(x_0^\star+s_\eta).
  \label{eq:equilibrium}
\end{equation}

\subsection{The small-weight displacement}
\label{app:small-weight}

Define
$F(s,\eta)=s+\eta J_R(x_0^\star+s)^{\!\top}R(x_0^\star+s)$.
Then $F(0,0)=0$ and $D_sF(0,0)=I$. If $R$ is sufficiently smooth, the
implicit function theorem gives a local solution $s_\eta$ near $\eta=0$.
Differentiating $F(s_\eta,\eta)=0$ at $\eta=0$ yields
\begin{equation}
  \left.\frac{d s_\eta}{d\eta}\right|_{\eta=0}
  =-J_R(x_0^\star)^{\!\top}R(x_0^\star),
\end{equation}
which gives Eq.~\ref{eq:population-shift}. For continuous $R$, global
minimizers exist by coercivity and approach $x_0^\star$ as $\eta\downarrow0$.
With local smoothness, they lie on this branch for small enough weight.

\subsection{The displacement bound}
\label{app:displacement-bound}

Since $x_\eta^\star$ minimizes the objective in
Eq.~\ref{eq:population-target}, its value cannot exceed the value at
$y=x_0^\star$. Substituting $s_\eta=x_\eta^\star-x_0^\star$ gives directly
\begin{equation}
  \|s_\eta\|^2
  \le \eta\left[\|R(x_0^\star)\|^2-\|R(x_\eta^\star)\|^2\right],
  \qquad
  \|s_\eta\|\le\sqrt\eta\,\|R(x_0^\star)\|.
  \label{eq:displacement-bound}
\end{equation}
The second inequality uses $\|R(x_\eta^\star)\|^2\ge0$. In particular,
$R(x_0^\star)=0$ implies $s_\eta=0$ for every $\eta\ge0$.

\subsection{The conditional residual gap}
\label{app:conditional-gap}

A nonzero residual at $x_0^\star$ is necessary for the combined objective
to displace the optimal denoising prediction, as shown by
Eq.~\ref{eq:displacement-bound}. This may seem unlikely when every clean
sample satisfies the PDE.
The optimal denoiser predicts $x_0^\star=\E[x_0\mid x_t]$, rather than
one of the feasible clean samples. It averages the clean samples compatible
with the noisy input $x_t$. Although each of these samples satisfies the
PDE, their average need not do so. This can happen when the PDE solution
set is not convex, since an average of solutions may lie outside it. If
$R$ is affine, conditional expectation preserves feasibility,
\[
  R(x_0^\star)=\E[R(x_0)\mid x_t]=0.
\]
For a nonlinear residual, feasibility of the conditional mean need not
follow. Let $\zeta=x_0-x_0^\star$. Taylor expansion of each residual
component around $x_0^\star$ gives
\begin{equation}
  R_k(x_0)=R_k(x_0^\star)+J_{R_k}(x_0^\star)\zeta
    +\tfrac12\zeta^{\!\top}\nabla^2R_k(x_0^\star)\zeta
    +r_k(\zeta),
\end{equation}
where $r_k(\zeta)=o(\|\zeta\|^2)$ as $\zeta\to0$. Provided the
conditional expectation of the remainder exists, $R_k(x_0)=0$ almost
surely and $\E[\zeta\mid x_t]=0$ give
\begin{equation}
  R_k(x_0^\star)
  =-\tfrac12\tr\!\left(\nabla^2R_k(x_0^\star)\Sigma_\post(x_t)\right)
    -\E[r_k(\zeta)\mid x_t].
  \label{eq:curvature}
\end{equation}
The remainder vanishes for quadratic $R$, making the identity exact.
Nonlinearity and posterior spread can therefore produce a nonzero residual
at the optimal denoising prediction, although neither guarantees it.

\paragraph{Neither ingredient is sufficient.} Let $R(x)=x_3-x_1^2+x_2^2$ with a
symmetric conditional law on $(a,a,0)$ and $(-a,-a,0)$, $a\neq0$. The mean is
feasible and both Hessian and covariance are nonzero, but their contraction
vanishes, so the displacement is zero at every weight despite a full-row-rank
Jacobian at the mean. The affine case is a sufficient exact control, not the only
zero-displacement case.

\paragraph{The circle identity and the origin.} For feasible circle data and
$R_0=\|x\|^2-1$, $R_0(x_0^\star)=-\tr\Sigma_\post$ exactly, and
$s_\eta=2\eta\tr\Sigma_\post\,x_0^\star+O(\eta^2)$, radially outward, partly
reversing the contraction caused by averaging distinct circle points, and
vanishing with the spread. At $x_0^\star=0$, however, $J_{R_0}(0)=0$ and the
objective $\tfrac12\|y\|^2+\tfrac\eta2(\|y\|^2-1)^2$ has minimizer $y=0$ for
$\eta\le1/2$, so nonzero residual must not be identified with nonzero first-order
displacement.

\subsection{An exact example of the trade-off}
\label{app:latitude}

Let
$\M=\{(a\cos\theta,a\sin\theta,h)\}$ with $a=\sqrt{\rho^2-h^2}$, $0<h<\rho$, on the
sphere of radius $\rho$, with a rotationally symmetric conditional law, so
$x_0^\star=(0,0,h)$. The residual depends only on $\|y\|$ while
$\|y-x_0^\star\|^2$ is minimized on the positive axis, so global minimization
reduces to $\min_{z\ge0}\tfrac12(z-h)^2+\tfrac\eta2(z^2-\rho^2)^2$, with derivative
$(z-h)+2\eta z(z^2-\rho^2)$, negative on $[0,h]$ and positive on
$[\rho,\infty)$. All minimizers lie in $(h,\rho)$, where stationarity is
\begin{equation}
  \eta(z)=\frac{z-h}{2z(\rho^2-z^2)},
  \qquad
  \eta'(z)=\frac{h(\rho^2-h^2)+(z-h)^2(2z+h)}{2z^2(\rho^2-z^2)^2}>0 ,
  \label{eq:latitude-monotone}
\end{equation}
increasing continuously from zero to infinity, so every positive weight has
exactly one stationary point in the interval, it is the global minimizer, and
$z_\eta$ increases from $h$ toward $\rho$. Every data point is equidistant from an
axis point, so
\begin{equation}
  |R(y_\eta)|=\rho^2-z_\eta^2\ \downarrow,
  \qquad
  \dist(y_\eta,\M)^2=\dist(x_0^\star,\M)^2+\|s_\eta\|^2\ \uparrow ,
  \label{eq:latitude}
\end{equation}
the added squared distance tending to $(\rho-h)^2$ as the residual tends to zero, so
every increment of weight improves the physics and takes the target farther
from every data state.

\subsection{The influence of the Jensen gap}
\label{sec:jensen}

For a nonlinear function the value at an average is not the average of the
values, and the difference is the Jensen gap. The gap has already been held responsible for the trade-off between physical and distributional accuracy, since the residual is imposed on the posterior mean rather than on a clean sample, and both \citet{zhang2026physicsinformed} and \citet{baldan2026physics} seek to reduce it by evaluating the residual closer to a clean sample. A denoiser makes such an average
unavoidable, since it sees only $x_t$ and must answer with one prediction for
every clean state compatible with it. Both terms in the objective
(Eq.~\ref{eq:objective}) then carry a gap. The denoising term carries the spread of
the clean samples compatible with $x_t$, the part $\tr\Sigma_\post$ of the
expected denoising loss that no prediction removes
(App.~\ref{app:conditional-loss}). The residual term can carry a nonzero residual
at their average $x_0^\star=\E[x_0\mid x_t]$ (Eq.~\ref{eq:curvature}), since each
sample is feasible but their average need not be. The two gaps differ in origin
but not in effect. Each one implies a nonzero prediction error
$\delta=\hat x_0-x_0$, and the comparison in Sec.~\ref{sec:ranking} depends on
the model only through the distribution of $\delta$, and once the residual cost
is linearized and the error has no preferred direction, only through
$\E\|\delta\|^2$.

That is the whole role of the gap. At a feasible clean sample a prediction with
zero error already minimizes both loss terms, so a nonzero error is needed before
they can favor different predictions. The gap is one source of such error, so
reducing it reduces both biases. Finite capacity, finite data and finite training
are others, and they persist when the averaging is removed. A perfect one-step
generator would carry no error and therefore no bias, while a real one is
imperfect for those reasons, which is all the comparison requires. The gap also
supplies no direction and the discussed biases are based on properties of the residual. 

\section{Derivation of the loss comparison}
\label{app:pairwise-derivation}

We derive Equation~\eqref{eq:pairwise} from the per-sample loss of
Equation~\eqref{eq:per_sample_loss}. Fix a clean training sample $x_0$ and a
noise level $t$, and write the loss of a prediction $\hat x_0$ as
\begin{equation*}
  \ell(\hat x_0)=\|\hat x_0-x_0\|^2+\eta\,\|R(\hat x_0)\|^2 ,
  \qquad \eta=c_t/\lambda_t .
\end{equation*}
As in Section~\ref{sec:ranking}, let $\delta$ be a random perturbation with
$\E[\delta]=0$ and $\E\|\delta\|^2<\infty$, and let $y\in\R^d$ be a second center.
We assume that $\E\|R(y+\delta)\|^2$ and
$\E\|R(x_0+\delta)\|^2$ are finite.

\paragraph{Step 1: Losses of the two predictions.} Substituting
$\hat x_0=y+\delta$ and $\hat x_0=x_0+\delta$ gives
\begin{align*}
  \ell(y+\delta)&=\|y-x_0+\delta\|^2+\eta\,\|R(y+\delta)\|^2,\\
  \ell(x_0+\delta)&=\|\delta\|^2+\eta\,\|R(x_0+\delta)\|^2 .
\end{align*}

\paragraph{Step 2: Expanding the denoising square.} Only the denoising term can
be expanded, since $R$ is a general nonlinear map evaluated at the prediction,
\begin{equation*}
  \|y-x_0+\delta\|^2=\|y-x_0\|^2+2\langle y-x_0,\delta\rangle+\|\delta\|^2 .
\end{equation*}
The residual term is kept as it stands and is treated in
Section~\ref{sec:sensitivity} by linearization.

\paragraph{Step 3: Taking expectations.} The vector $y-x_0$ is deterministic and
$\E[\delta]=0$, so the cross term vanishes,
\begin{equation*}
  \E\bigl[\langle y-x_0,\delta\rangle\bigr]=\langle y-x_0,\E[\delta]\rangle=0 .
\end{equation*}
The expected losses are therefore
\begin{align*}
  \E_\delta\bigl[\ell(y+\delta)\bigr]
    &=\|y-x_0\|^2+\E\|\delta\|^2+\eta\,\E\|R(y+\delta)\|^2,\\
  \E_\delta\bigl[\ell(x_0+\delta)\bigr]
    &=\E\|\delta\|^2+\eta\,\E\|R(x_0+\delta)\|^2 .
\end{align*}

\paragraph{Step 4: Difference of the expected losses.} The term $\E\|\delta\|^2$
appears in both expressions and is finite, so it cancels,
\begin{equation}
  \Delta(y)=\E_\delta\bigl[\ell(y+\delta)\bigr]
           -\E_\delta\bigl[\ell(x_0+\delta)\bigr]
  =\|y-x_0\|^2+\eta\,\E_\delta\!\left[
     \|R(y+\delta)\|^2-\|R(x_0+\delta)\|^2\right].
  \label{eq:pairwise-difference}
\end{equation}
The magnitude of the perturbation therefore does not enter the comparison, only the difference in expected residual between the two centers.

\paragraph{Step 5: The condition for a lower loss around $y$.} The prediction
around $y$ attains the lower expected loss if and only if $\Delta(y)<0$. Moving
the residual term to the other side gives
\begin{equation*}
  \eta\,\E_\delta\!\left[\|R(x_0+\delta)\|^2-\|R(y+\delta)\|^2\right]
  >\|y-x_0\|^2 ,
\end{equation*}
which is Equation~\eqref{eq:pairwise}.

\paragraph{Step 6: Why the condition can hold.} In contrast to the anchored case
of App.~\ref{app:anchored-relocation}, the left-hand side is not sign-constrained.
The residual is evaluated at the prediction, so the two centers are scored by the
residual geometry at two different places, and the expected residual cost around
$y$ can be smaller than the one around $x_0$. The derivation uses neither
$R(y)=0$ nor $R(x_0)=0$ and holds for every $y\in \Z$. Feasibility of the two
centers enters only in Section~\ref{sec:sensitivity}, where it removes the
zeroth-order term from the expansion of the residual.

\subsection{The centered-error assumption}
\label{app:centered}

The assumption $\E[\delta]=0$ enters at Step 3 alone. We repeat that step for a
general perturbation with mean $\bar\delta=\E[\delta]$. The cross term is now
$\E[\langle y-x_0,\delta\rangle]=\langle y-x_0,\bar\delta\rangle$, so
Equation~\eqref{eq:pairwise-difference} becomes
\begin{equation}
  \Delta(y)=\|y-x_0\|^2+2\langle y-x_0,\bar\delta\rangle
   +\eta\,\E_\delta\!\left[\|R(y+\delta)\|^2-\|R(x_0+\delta)\|^2\right],
  \label{eq:pairwise-full}
\end{equation}
and the condition for a lower loss around $y$ reads
\begin{equation*}
  \eta\,\E_\delta\!\left[\|R(x_0+\delta)\|^2-\|R(y+\delta)\|^2\right]
  >\|y-x_0\|^2+2\langle y-x_0,\bar\delta\rangle .
\end{equation*}

The right-hand side is no longer positive for every $y\neq x_0$. Completing the
square gives
$\|y-x_0\|^2+2\langle y-x_0,\bar\delta\rangle=\|y-(x_0-\bar\delta)\|^2-\|\bar\delta\|^2$,
so the condition is equivalent to
\begin{equation}
  \eta\,\E_\delta\!\left[\|R(x_0+\delta)\|^2-\|R(y+\delta)\|^2\right]
  >\|y-(x_0-\bar\delta)\|^2-\|\bar\delta\|^2 .
  \label{eq:pairwise-shifted}
\end{equation}
Two consequences follow. First, with $\bar\delta\neq0$ the center is no longer the mean of the prediction, since $y+\delta$ has mean $y+\bar\delta$. The comparison then ranks predictions whose means are $y+\bar\delta$ and $x_0+\bar\delta$, and the denoising term alone is minimized by the center that places the mean prediction on the target, $y=x_0-\bar\delta$, with value $-\|\bar\delta\|^2$. This is a matter of bookkeeping rather than a bias, because once the error carries a mean the decomposition of a prediction into a center and an error is no longer unique. Second, the displacement cost in Equation~\eqref{eq:pairwise-shifted} is then measured from $x_0-\bar\delta$ and is negative throughout the open ball of radius $\|\bar\delta\|$ around that point, so a center other than $x_0$ can appear preferable without the residual term contributing anything. Centering removes the ambiguity by making the center equal to the mean prediction, so that a preference for a center other than $x_0$ can come only from the residual term.

The same substitution applies to the anchored comparison. Repeating Step 3 of
App.~\ref{app:anchored-relocation} with $\E[\delta]=\bar\delta$ leaves
\begin{equation*}
  \Delta_{\mathrm{anch}}(y)=f(y-x_0+\bar\delta)-f(\bar\delta),
  \qquad
  f(u)=\|u\|^2+\frac{\eta}{N}\|Ju\|^2 ,
\end{equation*}
since both cross terms carry the factor $\bar\delta$. The function $f$ is a
strictly convex quadratic with $f(u)\ge\|u\|^2$, vanishing only at $u=0$, so
$\Delta_{\mathrm{anch}}$
is minimized uniquely at $y=x_0-\bar\delta$, the same center as for the denoising
term alone. The mean error therefore shifts the reference center of both
comparisons by the same amount and does not favor one objective over the other.
Centering fixes that reference at $x_0$, which is why it is assumed in
Section~\ref{sec:ranking} rather than carried through every expression.

\section{Derivation of the anchored comparison}
\label{app:anchored-relocation}

We derive Equation~\eqref{eq:anchored-relocation} from the anchored per-sample loss. Fix a clean training sample $x_0$ and write $J=J_R(x_0)$ and $N=\|J\|_F^2+\gamma$ with $\gamma>0$, so that $N>0$. The anchored loss of a prediction $\hat x_0$ is
\begin{equation*}
  \ell_{\mathrm{anch}}(\hat x_0)=\|\hat x_0-x_0\|^2+\frac{\eta}{N}\,\|J(\hat x_0-x_0)\|^2 .
\end{equation*}
Both $J$ and $N$ depend only on $x_0$, so they are the same for every prediction and do not depend on $\delta$. As in Section~\ref{sec:ranking}, let $\delta$ be a random perturbation with $\E[\delta]=0$ and $\E\|\delta\|^2<\infty$, let $y\in\R^d$ be an arbitrary center, and write the displacement of the center as $s=y-x_0$.

\paragraph{Step 1: Losses of the two predictions.} Substituting $\hat x_0=y+\delta$ and $\hat x_0=x_0+\delta$ gives
\begin{align*}
  \ell_{\mathrm{anch}}(y+\delta)&=\|s+\delta\|^2+\frac{\eta}{N}\,\|J(s+\delta)\|^2,\\
  \ell_{\mathrm{anch}}(x_0+\delta)&=\|\delta\|^2+\frac{\eta}{N}\,\|J\delta\|^2 .
\end{align*}

\paragraph{Step 2: Expanding the squares.} Expanding both squared norms in the first loss gives
\begin{align*}
  \|s+\delta\|^2&=\|s\|^2+2\langle s,\delta\rangle+\|\delta\|^2,\\
  \|J(s+\delta)\|^2&=\|Js\|^2+2\langle Js,J\delta\rangle+\|J\delta\|^2 .
\end{align*}

\paragraph{Step 3: Taking expectations.} All expectations are finite, since $\E\|J\delta\|^2\le\|J\|_F^2\,\E\|\delta\|^2<\infty$ by the Cauchy--Schwarz inequality. Because $s$ and $J$ are deterministic and $\E[\delta]=0$, both cross terms vanish:
\begin{align*}
  \E\bigl[\langle s,\delta\rangle\bigr]&=\langle s,\E[\delta]\rangle=0,\\
  \E\bigl[\langle Js,J\delta\rangle\bigr]&=\langle J^{\top}Js,\E[\delta]\rangle=0 .
\end{align*}
The expected losses are therefore
\begin{align*}
  \E_\delta\bigl[\ell_{\mathrm{anch}}(y+\delta)\bigr]&=\|s\|^2+\E\|\delta\|^2+\frac{\eta}{N}\Bigl(\|Js\|^2+\E\|J\delta\|^2\Bigr),\\
  \E_\delta\bigl[\ell_{\mathrm{anch}}(x_0+\delta)\bigr]&=\E\|\delta\|^2+\frac{\eta}{N}\,\E\|J\delta\|^2 .
\end{align*}

\paragraph{Step 4: Difference of the expected losses.} The terms $\E\|\delta\|^2$ and $\frac{\eta}{N}\E\|J\delta\|^2$ appear in both expressions and are finite, so they cancel in the difference:
\begin{equation*}
  \Delta(y)=\E_\delta\bigl[\ell_{\mathrm{anch}}(y+\delta)\bigr]-\E_\delta\bigl[\ell_{\mathrm{anch}}(x_0+\delta)\bigr]=\|y-x_0\|^2+\frac{\eta}{N}\,\|J(y-x_0)\|^2 .
\end{equation*}

\paragraph{Step 5: The condition for a lower loss around $y$.} The prediction around $y$ attains the lower expected loss if and only if $\Delta(y)<0$. Moving the anchored term to the other side gives
\begin{equation*}
  -\eta\,\frac{\|J_R(x_0)(y-x_0)\|^2}{\|J_R(x_0)\|_F^2+\gamma}>\|y-x_0\|^2,
\end{equation*}
which is Equation~\eqref{eq:anchored-relocation}.

\paragraph{Step 6: The condition never holds.} For $\eta\ge 0$, the left-hand side is at most zero, because it is the negative of a squared norm divided by $N>0$. For $y\neq x_0$, the right-hand side is strictly positive. The condition therefore fails for every $y\neq x_0$, and equivalently $\Delta(y)>0$ for all $y\neq x_0$. Hence $x_0$ is the unique minimizer of $y\mapsto\E_\delta[\ell_{\mathrm{anch}}(y+\delta)]$. The derivation uses neither $R(y)=0$ nor any linearization of $R$, so the result holds for every $y\in\R^d$ and is exact.

\section{The conditional target of the anchored loss}
\label{app:anchored}

Write
\begin{equation*}
  B(x_0):=\frac{J_R(x_0)^{\!\top}J_R(x_0)}{\|J_R(x_0)\|_F^2+\gamma},
  \label{eq:B}
\end{equation*}
so that the residual term of Eq.~\ref{eq:anchored-objective} is
$\eta\,\delta^{\!\top}B(x_0)\delta$ with $\delta=\hat x_0-x_0$. For a fixed clean
target the combined loss is minimized uniquely at $\hat x_0=x_0$, since its
gradient is $2\lambda_t\left[I+\eta B(x_0)\right](\hat x_0-x_0)$ and the matrix
is positive definite.

The comparison in Sec.~\ref{sec:anchoring} fixes a clean target $x_0$. The
network, however, receives only the noisy input $x_t$ and the noise level $t$.
Several clean samples may be compatible with that input, so the best prediction
minimizes the loss averaged over their conditional distribution. For ordinary
squared-error denoising this prediction is the conditional mean
$x_0^\star=\E[x_0\mid x_t,t]$. Anchoring can change this optimum because
different clean samples may have different weighting matrices $B(x_0)$.

To see this, fix $(x_t,t)$ and let $y\in\R^d$ denote the network prediction.
Assuming finite conditional second moments, its expected anchored loss is
\begin{equation}
  \E\!\left[
    (y-x_0)^{\!\top}\left[I+\eta B(x_0)\right](y-x_0)
    \mid x_t,t
  \right].
  \label{eq:anchored-conditional}
\end{equation}
At the optimum the average weighted prediction error must be zero.
Differentiating with respect to $y$ and rearranging gives
\begin{equation}
  \bigl[I+\eta\,\E[B(x_0)\mid x_t,t]\bigr](y_{\mathrm{anch}}^\star-x_0^\star)
  =\eta\,\E[B(x_0)(x_0-x_0^\star)\mid x_t,t].
  \label{eq:anchored-target}
\end{equation}
The matrix on the left is positive definite, so this equation uniquely
determines the shift from the ordinary denoising target.

For $\eta>0$ the optimum remains $x_0^\star$ exactly when
\[
  \E[B(x_0)(x_0-x_0^\star)\mid x_t,t]=0.
\]
The deviations $x_0-x_0^\star$ average to zero by definition. The condition
requires that they still average to zero after each is weighted by its sample's
matrix $B(x_0)$. A sufficient condition is that $B(x_0)$ is the same for all
clean samples in the conditional support. It can then be taken outside the
expectation, leaving $\E[x_0-x_0^\star\mid x_t,t]=0$. This is what happens for a
fixed affine residual $R(x)=Ax-b$, where $J_R\equiv A$ and, at a feasible $x_0$,
the anchored residual equals the original one, $J_R(x_0)(\hat x_0-x_0)=R(\hat
x_0)$, so $B$ does not depend on the sample. If the weighting varies across samples, the
cancellation may fail and the optimal prediction can move away from $x_0^\star$.

For the latitude example of App.~\ref{app:latitude} with $\gamma=0$,
$B=x_0x_0^{\!\top}/\rho^2$ and rotational symmetry give
$(y_{\mathrm{anch}}^\star)_3=h(1+\eta)/(1+\eta h^2/\rho^2)>h$, so even the
normalized anchored loss can increase the distance to the data in this
conditional example.

Anchoring therefore removes the preference for moving toward another solution in
the centered, fixed-perturbation comparison. It does not generally preserve the
ordinary denoising optimum after averaging over possible clean targets.

\FloatBarrier
\section{Additional related work}
\label{app:related_work}

\paragraph{Generative models for physical systems.} Diffusion and flow matching models are increasingly used to learn distributions over physical states, with applications ranging from turbulent and aerodynamic flows \citep{li2024synthetic,liu2024uncertainty} and inverse PDE problems \citep{holzschuh2023solving} to particle physics \citep{mikuni2023fast} and the generation of molecules and proteins \citep{schneuing2024structure,bose2024se,darouich2026adaptive}. Unlike deterministic surrogates, they represent the full distribution of solutions, which makes them suited to uncertainty quantification and ensemble generation. The idea of enforcing physics through the training objective comes from physics-informed neural networks \citep{raissi2019physics}, which add the PDE residual to the loss of a deterministic model, and whose optimization is known to suffer from conflicts between the loss terms \citep{krishnapriyan2021characterizing,liu2025config}.

\paragraph{The trade-off between physics and distribution.}
\citet{baldan2026physics} state that trade-off explicitly and report it as a Pareto front. They attribute part of it to the Jensen gap, which they narrow by unrolling several steps before the residual is evaluated, and they resolve the remaining disagreement between the two objectives at the level of the gradients, with a conflict-free update \citep{liu2025config}. We locate that conflict and the resulting displacement in the residual term itself and change the loss, anchoring the residual at the clean training sample, which removes part of the conflict instead of balancing it. Also \citet{yeom2025softly} argue that a residual regulariser breaks the distributional guarantees of the model and move the constraint into the forward pass instead. \citet{chang2026stabilizing} also observe that the residual term can lead to a collapse onto smooth samples that are less sensitive under the residual, and they respond with a training scheme rather than with a different penalty, learning the distribution first and fine-tuning it under the residual afterwards, with the decoder that produces the coefficient field frozen so that the residual cannot change it. \citet{yuan2025pirf} call the trade-off reward hacking and report a fine-tuned model that reaches the lowest residual and the most distorted field, which they contain by updating only part of the network and penalizing the drift of the weights away from the pretrained ones. \citet{tauberschmidt2026physics} steer a pretrained model toward the residual-tilted distribution and anchor it to the pretrained one, and report the same trade-off along that anchor, although their base model is trained on low-fidelity data, so for them the departure is intended. 
\citet{xu2026right} also relate the distortion to the sensitivity of the constraint, but for inference methods that enforce a hard constraint post hoc on a fixed pretrained model by projection or guidance and thereby omit a co-area factor $[\det JJ^\top]^{-1/2}$. The works above and ours instead place the residual in the training objective.
Prior work establishes that physical residual objectives
can distort a learned distribution and motivates methods that restrict or balance their influence.
Our contribution is to characterize sensitivity-dependent
relocation through a controlled, fixed-perturbation comparison, distinguish feasibility from membership in the data support, and construct a target-anchored quadratic
penalty that removes the relocation preference in this
comparison. We also characterize the conditional-target
shift that can remain after anchoring.

\FloatBarrier
\section{Experimental setup}
\label{app:setup}

\subsection{Datasets}
\label{app:datasets}

\subsubsection{The shocked circle}
\label{app:circle}

\begin{figure}[!htbp]
\centering
\includegraphics[width=0.6\linewidth]{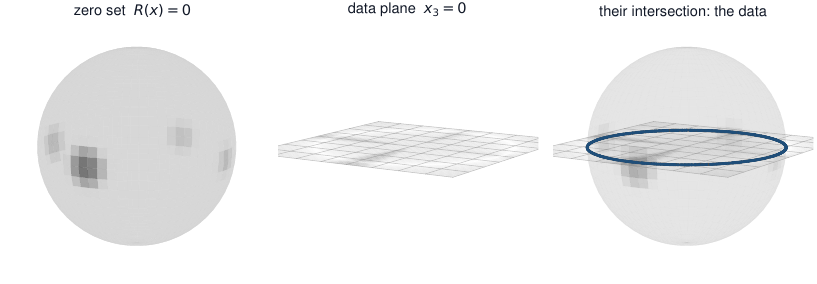}
\caption{Data support and zero set of the residual, drawn in three dimensions for
visualization. The residual is satisfied on the whole sphere, and the data are only
the circle where the sphere meets the data plane. The experiments use $D=10$.}
\label{fig:datavssolution}
\end{figure}

\paragraph{What the dataset is for.} We identify two conditions for the described destributional biases, (1) a residual
cost that varies along the data and (2) a solution set larger than the data support.
Both are properties of the pair $(R,\pdata)$. On a PDE they are hard to read off
and cannot be varied one at a time. This dataset makes both explicit. The data are
a circle, and the residual is satisfied on the whole sphere that contains this
circle. A positive factor $g$ multiplying the residual changes the penalty an error
incurs without changing a single feasible state, as visualized in
Fig.~\ref{fig:datavssolution}.

\begin{figure}[!htbp]
\centering
\includegraphics[width=0.62\linewidth]{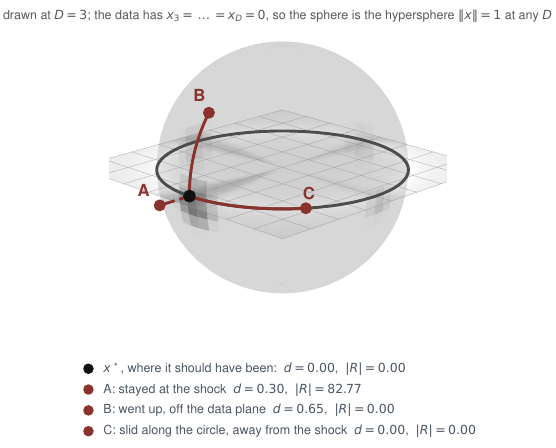}
\caption{The three things a penalized prediction can do, drawn in three dimensions.
In black, where the sample should be, at a state where the residual is sensitive.
\textbf{(A)} Correct prediction. \textbf{(B)} Leakage, which leaves the data plane and therefore the data support along the zero set, where $|R|=0$. \textbf{(C)} Avoidance, where the prediction moves to a region of lower stiffness, which may or may not lie on the data support.}
\label{fig:schematic-app}
\end{figure}

Figure~\ref{fig:schematic-app} summarizes the three cases. A prediction that
belongs at a state where the residual is stiff can do three things. It can predict
the sample as accurately as it is able and incur the penalty its error attracts
there, which is the intended behaviour. It can move to a state at which the same
error is penalized less, which is avoidance, and that state may lie on the data
circle or off it. Or it can leave the data plane along the sphere, which reaches
zero residual at a state the data never produce, and that is leakage. A penalty
built from $R(\hat x_0)$ alone cannot distinguish either from success, because
both land on $\Z$.

\paragraph{Data and residual.} The data manifold is the unit circle in the first
two coordinates of $\R^D$,
\begin{equation}
  \M=\bigl\{x\in\R^D:\;x_1^2+x_2^2=1,\;x_3=\dots=x_D=0\bigr\},
  \qquad D=10.
  \label{eq:toy-data}
\end{equation}
Samples are $x=(\cos\theta,\sin\theta,0,\dots,0)$ with $\theta\sim\mathcal U[0,2\pi)$.
The training set contains $20\,000$ such points. The residual is
\begin{equation}
  R(x)=\bigl(\|x\|^2-1\bigr)\,g(x),
  \label{eq:toy-residual}
\end{equation}
a scalar, so $m=1$. Since $g\ge1>0$, the zero set is $\Z=\mathbb S^{D-1}$. Hence
$\M=\Z\cap\{x_3=\dots=x_D=0\}$ and $\dim\Z-\dim\M=D-2=8$. Eight directions lead
from the data along the zero set, and the residual cannot see a sample drifting
along any of them.

\begin{figure}[!htbp]
\centering
\includegraphics[width=0.5\linewidth]{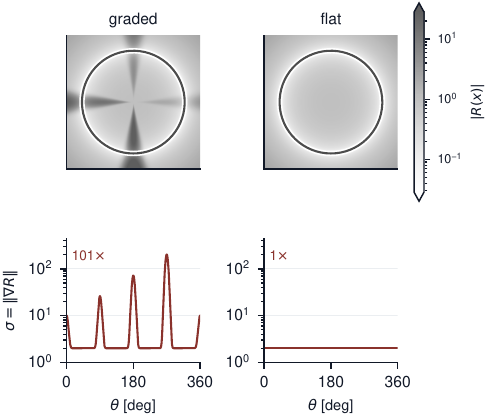}
\caption{One data circle, two residuals. Top, the residual in the data plane, where the zero set is the same circle for both. Bottom, the residual cost $2g$ along the data, which spans a factor of $101$ for \texttt{graded} and is constant for \texttt{flat}.}
\label{fig:datasets}
\end{figure}

\paragraph{Why the stiffness is the residual cost.} $R$ vanishes on all of $\Z$,
which contains $\M$, so the derivative of $R$ along the data vanishes and $\nabla R$
is purely radial. At a data point the factor $\|x\|^2-1$ eliminates the $\nabla g$ term.
This leaves $\nabla R(x_0)=2g(x_0)x_0$ and hence
$\sigma(\theta)=2g(\theta)/\sqrt D$ exactly, which we verify numerically to
$3\times10^{-5}$. An error of size $\varepsilon$ made at angle $\theta$ therefore
produces a residual $|R|\approx2g(\theta)\varepsilon$. The stiffness profile is not a
proxy for the cost. It is the cost itself, and it can be read directly from
Fig.~\ref{fig:datasets}.

\paragraph{The stiffness.} In Eq.~\ref{eq:toy-stiffness}, $g$ is a sum of four
von Mises-Fisher bumps on the direction $\hat x=x/\|x\|$,
\begin{equation}
  g(x)=1+m(\|x\|)\sum_{k=1}^{4}h_k
       \exp\!\Bigl(-\tfrac{2(1-\hat x\cdot c_k)}{w_k^2}\Bigr),
  \qquad (c_k)=(+e_1,+e_2,-e_1,-e_2),
  \label{eq:toy-stiffness}
\end{equation}
so the $k$-th center meets the data circle at $\theta_k=(k-1)\pi/2$. Since
$2(1-\cos\gamma)=\gamma^2+O(\gamma^4)$ for the geodesic angle $\gamma$, each bump is a Gaussian of angular width $w_k$ to leading order. Writing it in the dot product rather than in $\arccos$ keeps $\nabla R$ finite at $\hat x=\pm c_k$. The radial mask $m(r)=(1+a^2)r^2/(r^2+a^2)$ with $a=0.8$ satisfies $m(1)=1$ and $m(0)=0$. It removes a tangential singularity that $\nabla R$ would otherwise carry at the origin. Without it we measure $\|\nabla R\|\approx1.3\times10^5$ at $\|x\|=2\times10^{-3}$. This matters because the $x_0$-prediction lies near the origin at high noise. Since
$m(1)=1$, the stiffness on $\M$ is unchanged. Each bump is a localized spherical cap rather than a wedge constant along $x_3,\dots,x_D$.

\paragraph{The two residuals.} Both share the data and differ only in $g$, so both have the same feasible set (Fig.~\ref{fig:datasets}). \texttt{graded} places four sensitive regions of increasing strength on the circle. \texttt{flat} sets $g\equiv1$ and is the plain hypersphere residual $R(x)=\|x\|^2-1$. Table~\ref{tab:toy-profiles} lists the parameters.

\begin{table}[!htbp]
\centering\small
\setlength{\tabcolsep}{6pt}
\caption{The two residual profiles of the shocked circle.}
\label{tab:toy-profiles}
\begin{tabular}{lcc}
\toprule
& \texttt{graded} & \texttt{flat} (control)\\
\midrule
heights $h_k$ & $4,12,35,100$ & ---\\
widths $w_k$  & $0.10$ & ---\\
$2g$ along $\M$ & $2\!\to\!202$ ($101\times$) & $2$ ($1\times$)\\
bump HWHM & $4.8^\circ$ & ---\\
\bottomrule
\end{tabular}
\end{table}

\subsubsection{PDE datasets}
\label{app:pde-data}
Both datasets are generated with \texttt{torchfsm} \citep{liu_2025_15350210} on
$\xi\in[0,1)$ with periodic boundary conditions and generator seed $42$. Each
contains $10\,000$ trajectories, split $90/10$ into training and held-out data with
a fixed seed.

\paragraph{Advection.} $\beta=1$, $N_\xi=256$, $\Delta\tau=1/256$ and $255$ steps,
so each sample is a $256\times256$ field spanning $\tau\in[0,255/256]$. The initial
conditions are
$u_0=\sum_{i=1}^2A_i\sin(2\pi\ell_i\xi/L+\varphi_i)$ with
$\ell_i\sim\mathcal U\{1,\dots,5\}$, $A_i\sim\mathcal U[0,1)$ and
$\varphi_i\sim\mathcal U[0,2\pi)$. With probability $0.1$ the initial condition is
replaced by $\pm|u_0|$ with a random sign, and with probability $0.1$ it is
multiplied by $\sin^2(\pi\xi/L)$.

\paragraph{Burgers.} $\nu=10^{-3}$, $N_\xi=512$, $\Delta\tau=5\times10^{-4}$ and
$255$ steps, so each sample is a $256\times512$ field spanning
$\tau\in[0,0.1275]$. The initial conditions are
$\tilde u_0=\sum_{i=1}^4\frac{a_i}{\max(1,f_i)}\sin(2\pi f_i\xi/L+\varphi_i)$ with
$f_i\sim\mathcal U\{1,\dots,4\}$ and $a_i\sim\mathcal N(0,1)$. They are made mean-free
and rescaled to $u_0=\alpha(\tilde u_0-\overline{\tilde u_0})/
\|\tilde u_0-\overline{\tilde u_0}\|_\infty$ with $\alpha\sim\mathcal U[0.8,1.2]$.
Every trajectory thus starts with comparable energy and zero mean, and the periodic
dynamics conserve the mean.

\paragraph{Darcy.} The domain is the unit square with $N=64$ grid points per axis, including
the boundary, so $h=1/63$ and each sample is a pair of $64\times64$ fields $(p,K)$. The
log-permeability is a Gaussian random field with covariance $\exp(-\|x-x'\|/\ell)$, $\ell=0.1$,
represented by its first $64$ Karhunen--Lo\`eve modes with standard normal coefficients, and
$K=\exp(\log K)$. The source is $f=10$ on $[0,0.125]^2$ and $f=-10$ on $[0.875,1]^2$, and zero
elsewhere. For each $K$, the pressure solves $-\nabla\cdot(K\nabla p)=f$ with zero-flux
boundaries and zero mean, discretized with second-order finite differences and solved in the
least-squares sense. We use the dataset of
\citet{bastek2025physics}, which contains $10\,000$ training and $1\,000$ held-out samples.

\paragraph{Residuals.} The residuals are evaluated on the generated field
$u(\xi,\tau)$. Time derivatives use second-order finite-difference stencils along
the non-periodic time axis. For advection,
$R=\partial_\tau u+\beta\partial_\xi u$ with a second-order periodic spatial
stencil. For Burgers, the spatial operators are spectral, consistent with the
generating solver,
\[
  R=\partial_\tau u+\mathcal D[u]\,\partial_\xi\mathcal D[u]-\nu\partial_{\xi\xi}u,
\]
where $\mathcal D$ is the $2/3$ dealiasing filter. It is applied to the quadratic
term only, so that aliasing errors do not appear as spurious residual. One-sided
stencils near the boundary are excluded from the residual. For Burgers these are the
first and last five time rows. For advection, the first and last five rows in time
and in space are excluded.
For Darcy, the residual is the strong form
\[
  R=-K\Delta p-\nabla K\cdot\nabla p-f
\]
with second-order finite-difference stencils.

\FloatBarrier
\subsection{Metrics}
\label{app:metrics}

Let $\{\hat x^{(i)}\}_{i=1}^{N}$ be the generated samples and
$\{x^{(j)}\}_{j=1}^{N_{\mathrm{ref}}}$ the reference data samples.

\paragraph{Mean residual.} With $\mathcal I\subseteq\{1,\dots,m\}$ the residual
entries not excluded at the boundary (App.~\ref{app:pde-data}),
\begin{equation}
  \overline{|R|}=\frac1N\sum_{i=1}^{N}\frac{1}{|\mathcal I|}
  \sum_{k\in\mathcal I}\bigl|R_k(\hat x^{(i)})\bigr|.
  \label{eq:metric-residual}
\end{equation}

\paragraph{Maximum mean discrepancy.} Each field $x$ is mapped to a vector of $q$ per-sample
features $\phi(x)\in\R^q$, standardized with the mean $\mu$ and standard deviation $s$ of the
features of the reference data, $z=(\phi(x)-\mu)/s$. With the Gaussian kernel
$k(z,z')=\exp\bigl(-\|z-z'\|^2/(2h^2)\bigr)$, where $h$ is the median distance between all pairs
of the pooled generated and reference features, we report the unbiased estimate
\citep{gretton2012kernel}
\begin{equation}
\begin{aligned}
  \mathrm{MMD}^2={}&\frac{1}{N(N-1)}\sum_{i\neq i'}k(\hat z_i,\hat z_{i'})
  +\frac{1}{N_{\mathrm{ref}}(N_{\mathrm{ref}}-1)}\sum_{j\neq j'}k(z_j,z_{j'})\\
  &-\frac{2}{N N_{\mathrm{ref}}}\sum_{i,j}k(\hat z_i,z_j),
\end{aligned}
\label{eq:metric-mmd}
\end{equation}
set to zero if negative, with $N=N_{\mathrm{ref}}=1000$. On Advection and Burgers the features
are $\max|\partial_\xi u|$, the total variation $\sum|u_{k+1}-u_k|$ along $\xi$ summed over
$\tau$, $\mathrm{std}(u)$ and $\max|u|$. On Darcy they are $\mathrm{std}(p)$, $\max|\nabla p|$,
and the mean, standard deviation and correlation at a lag of $0.1$ of $\log K$.

\paragraph{Wasserstein distance.} For two distributions $P$ and $Q$ on $\R$ with
cumulative distribution functions $F_P$ and $F_Q$ \citep{villani2009optimal, scipy2025},
\begin{equation}
  W_1(P,Q)=\int_{\R}\bigl|F_P(s)-F_Q(s)\bigr|\,ds .
  \label{eq:metric-wd}
\end{equation}

\paragraph{Forward error.} For darcy dataset, for a generated pair $(p,K)$ we solve
$-\nabla\cdot(K\nabla p^{*})=f$ with zero-flux boundaries, using the finite-difference system of
the data generator, and report the median over samples of
$\|p-\bar p-p^{*}(K)\|_2/\|p^{*}(K)\|_2$, where $\bar p$ is the mean of $p$. The pressure is
defined only up to a constant under these boundary conditions, so the mean is removed. On the
held-out data the forward error is $3.4\cdot10^{-5}$, the precision of the stored fields.

\paragraph{Sensitivity ratio.} The residual sensitivity of a state $x$ is the
state-dependent part of the residual Jacobian,
\begin{equation}
  \sigma(x)=\frac{\|J_R(x)-J_R(0)\|_F}{\sqrt m},
  \label{eq:sigma}
\end{equation}
where $\|A\|_F^2=\E_v\|Av\|^2$, $v\sim\mathcal N(0,I)$, is estimated with two
Jacobian--vector products \citep{hutchinson1989stochastic}.
and the sensitivity ratio is
\begin{equation}
  \frac{\operatorname{median}_i\,\sigma(\hat x^{(i)})}
       {\operatorname{median}_j\,\sigma(x^{(j)})}.
  \label{eq:metric-sigma}
\end{equation}

\paragraph{Amplitude ratio.} With
$\mathrm{std}(x)=\bigl(\frac1d\sum_{k=1}^{d}(x_k-\bar x)^2\bigr)^{1/2}$ and
$\bar x=\frac1d\sum_k x_k$,
\begin{equation}
  \frac{\mathrm{std}(u)_{\mathrm{gen}}}{\mathrm{std}(u)_{\mathrm{data}}}
  =\frac{\frac1N\sum_i\mathrm{std}(\hat x^{(i)})}
        {\frac1{N_{\mathrm{ref}}}\sum_j\mathrm{std}(x^{(j)})}.
  \label{eq:metric-amplitude}
\end{equation}

\paragraph{Distance to the data plane.} For $x\in\R^D$,
\begin{equation}
  \|x_{3:D}\|=\Bigl(\sum_{k=3}^{D}x_k^2\Bigr)^{1/2},
  \label{eq:metric-offplane}
\end{equation}
reported as the median over the generated samples.

\paragraph{Transition-state metrics.} Each generated structure $x$ with atomic positions $r_1,\dots,r_N$ is compared with the reference TS $x^{\mathrm{ref}}$ of the same reaction. The root mean square deviation measures differences in atomic positions,
\begin{equation*} \mathrm{RMSD}=\Bigl(\tfrac{1}{N}\textstyle\sum_{i=1}^{N}\|r_i-r_i^{\mathrm{ref}}\|^2\Bigr)^{1/2}, \end{equation*}
computed after optimal rigid alignment of the two structures. The distance matrix absolute error measures the mean absolute difference in interatomic distances $d_{ij}=\|r_i-r_j\|$,
\begin{equation*} \mathrm{DMAE}=\tfrac{2}{N(N-1)}\textstyle\sum_{i<j}\bigl|d_{ij}-d_{ij}^{\mathrm{ref}}\bigr|, \end{equation*}
and is invariant to rotations and translations. The energy difference $\Delta E=|E(x)-E(x^{\mathrm{ref}})|$, evaluated at the GFN2-xTB level, measures energetic agreement. The maximum force $F_{\max}=\max_i\|F_i(x)\|$, with $F=-\nabla E$, measures the departure from stationarity on the potential energy surface (PES). The Hessian MAE, the mean absolute difference between the entries of the generated and reference Hessians, measures agreement in the local curvature of the PES. The valid TS rate is the percentage of generated structures whose Hessian has exactly one negative eigenvalue, as required for a first-order saddle point.

\FloatBarrier
\subsection{Implementation details}
\label{app:implementation}

For each system, DM, PIDM and the anchored objective share the same denoiser,
optimizer, noise schedule and training budget, and differ only in the physics term.
The unpenalized baseline, denoted DM, sets $c=0$ in Eq.~\eqref{eq:objective}. All
models use the clean-sample parameterization, so $\hat x_0$ is the network output.
The noise level $t$ is drawn uniformly, and the cosine variance schedule is used for
$\{\beta_t\}$. The penalty is applied to every sample at every training step,
without thresholding or gating. Table~\ref{tab:hyperparameters} lists the settings.

\paragraph{Weights of the two loss terms.} The denoising weight is
$\lambda_t=\mathrm{SNR}(t)=\bar\alpha_t/(1-\bar\alpha_t)$, which makes the
denoising term equal to the standard noise-prediction loss. On Advection and
Burgers it is clipped, $\lambda_t=\min(\mathrm{SNR}(t),5)$. For PIDM, the residual
weight is the schedule of \citet{bastek2025physics},
$\omega_t=1/(2\tilde\beta_t)$, where $\tilde\beta_t$ is the posterior variance of the
forward process and its value at $t=0$ is replaced by that at $t=1$. The anchored
objective uses the denoising weight for both terms (Eq.~\eqref{eq:anchored-objective}),
so the ratio of the two terms is $\eta=c$ at every noise level.

\paragraph{Residual terms.} On Advection and Burgers, both residual terms are
averaged over the $m$ residual entries. PIDM uses $\frac1m\|R(\hat x_0)\|^2$. The
anchored term $J_R(x_0)\delta$ is computed with one forward-mode Jacobian--vector
product, with $x_0$ detached. The normalization $\|J_R(x_0)\|_F^2$ is estimated
with Hutchinson's estimator using one Gaussian probe per sample and step, and it is
excluded from the gradient. The constant in the denominator is $\gamma=10^{-12}$.
On the shocked circle the residual is scalar, so $J_R(x_0)=\nabla R(x_0)^\top$ and
its norm are computed exactly.

\paragraph{Training.} We train with Adam, without weight decay or learning-rate
schedule, and clip the gradient norm at $1$. Every model is trained for
$100\,000$ iterations and evaluated at its last checkpoint, without early stopping.
On Advection, Darcy and Burgers, $256$ of the $9000$ training fields are held out as a
stopping set and not used for training, so each model sees $8744$ fields. 

\paragraph{Sampling.} Samples are generated with ancestral DDPM sampling over
the full schedule, without noise in the last step.

\begin{table}[!htbp]
\centering\small
\setlength{\tabcolsep}{6pt}
\caption{Training and sampling settings. The same settings are used for DM, PIDM
and the anchored objective.}
\label{tab:hyperparameters}
\begin{tabular}{lccc}
\toprule
& Shocked circle & Advection and Burgers & Darcy \\
\midrule
Denoiser & MLP & U-Net & U-Net \\
Architecture & $5\times512$, SiLU & channels $32,64,128,256$ & as Burgers \\
& & 2 ResNet blocks per level, & \\
& & linear attention per level, & \\
& & attention in the bottleneck & \\
Input channels & -- & $1$ ($u$) & $2$ ($p$, $K$) \\
Time input & $t/T$, repeated to $64$ dims & sinusoidal, then MLP ($128$) & sinusoidal, then MLP ($128$) \\
Trainable parameters & $1.09$M & $8.9$M & $8.9$M \\
Diffusion steps $T$ & $1000$ & $100$ & $100$ \\
Variance schedule & cosine & cosine & cosine \\
Denoising weight $\lambda_t$ & $\mathrm{SNR}(t)$ & $\min(\mathrm{SNR}(t),5)$ & $\min(\mathrm{SNR}(t),5)$ \\
Optimizer & Adam & Adam & Adam \\
Learning rate & $10^{-3}$ & $10^{-4}$ & $10^{-4}$ \\
Batch size & $512$ & $8$ & $8$ \\
Iterations & $100\,000$ & $100\,000$ & $100\,000$ \\
Gradient clipping & $1.0$ & $1.0$ & $1.0$ \\
EMA decay & $0.999$ & $0.999$ (Advection), none (Burgers) & none \\
Training samples & $20\,000$ & $8744$ & $10\,000$ \\
Sampler & DDPM, $1000$ steps & DDPM, $100$ steps & DDPM, $100$ steps \\
Generated samples & $40\,000$ & $1000$ & $1000$ \\
Seed & $0$ & $0$ & $0$ \\
GPU & RTX 4090 & A40 & A40 \\
\bottomrule
\end{tabular}
\end{table}

\paragraph{Transition state generation.} We follow the procedure of \citet{darouich2026curvature}. As the surrogate potential, we use the EquiformerV2 HIP checkpoint \citep{burger2025shoot} fine-tuned on GFN2-xTB Hessians of Transition1x-2p3p4p, which reaches a Hessian MAE of $3.51$\,kcal/mol/\AA$^2$ on this dataset. For TS generation, we start from the adapted OA-ReactDiff checkpoint of \citet{darouich2026robust}, which covers elements beyond H, C, N, and O, and fine-tune it for 200 epochs with the React-OT flow matching loss at a learning rate of $2.5\times10^{-4}$. From this shared starting point, FM, FM + residual, and FM + anchored are each trained for a further 200 epochs at the same learning rate, with a batch size of 16 on a single A100 GPU. Following \citet{darouich2026curvature}, both residual variants minimize
\begin{equation*} \mathcal{L}(\theta)=\mathcal{L}_{\mathrm{FM}}+\eta\,t^{p}\,\mathcal{R}(\hat x_1,x_1), \qquad \hat x_1=x_t+(1-t)\,u_\theta(x_t,t), \end{equation*}
where $x_1$ is the reference TS, $\hat x_1$ the single-step estimate of the clean endpoint, and the weight $t^{p}$ down-weights the imprecise estimates obtained at high noise. Note that flow matching places the clean sample at $t=1$, so $x_1$ here corresponds to $x_0$ in the main text. FM + residual uses the force predicted by HIP, $\mathcal{R}=\|F_\phi(\hat x_1)\|^2$, and FM + anchored uses the anchored penalty with the HIP Hessian at the reference TS, $\mathcal{R}=\|H_\phi(x_1)(\hat x_1-x_1)\|^2/(\|H_\phi(x_1)\|_F^2+\gamma)$, which is precomputed once per training structure. For both variants, the weight $\eta$ and the exponent $p$ are optimized on the validation set using Bayesian optimization.

\FloatBarrier
\subsection{Training cost}
\label{app:cost}

Table~\ref{tab:cost} compares the training time of the three objectives. PIDM
evaluates the residual at the prediction and differentiates through it. The
anchored objective additionally needs one Jacobian--vector product at $x_0$ for
$J_R(x_0)\delta$ and one for the estimate of $\|J_R(x_0)\|_F^2$.

\begin{table}[!htbp]
\centering\small
\setlength{\tabcolsep}{6pt}
\caption{Wall-clock time in seconds per $1000$ training iterations on one A40 GPU at batch size
$8$, for the training step alone. In parentheses the increase over DM.}
\label{tab:cost}
\begin{tabular}{lccc}
\toprule
& DM & PIDM & Anchored \\
\midrule
Advection & $526$ & $531$ ($+1.0\%$) & $539$ ($+2.5\%$) \\
Burgers & $1067$ & $1071$ ($+0.4\%$) & $1082$ ($+1.5\%$) \\
Darcy & $41$ & $57$ ($+38\%$) & $92$ ($+123\%$) \\
\bottomrule
\end{tabular}
\end{table}
On Advection and Burgers the anchored objective takes $2.5\%$ and $1.5\%$ more time per
iteration than DM, and PIDM takes $1.0\%$ and $0.4\%$ more. There the step is dominated by the
forward and backward pass of the U-Net, so the two additional Jacobian--vector products add
little.
Darcy has the smallest fields and the
cheapest step, so the residual is a large share of it. There PIDM takes $38\%$ more time than DM
and the anchored objective $123\%$ more, which is still less than a tenth of a Burgers step in
absolute terms.

\FloatBarrier
\section{Additional results}
\label{app:results}

\subsection{Shocked circle}
\label{app:toy_detail}
\begin{figure*}[!t]
\centering
\includegraphics[width=\textwidth]{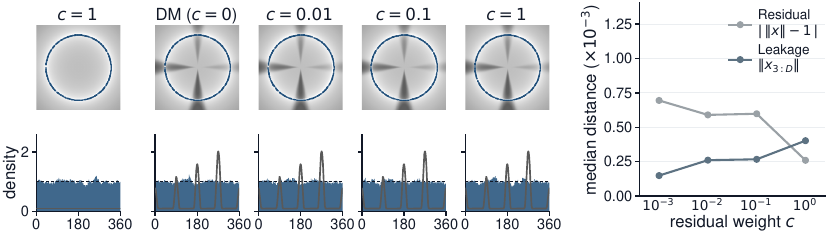}
\caption{Anchored on the shocked circle at $D=10$. Top, samples in the data plane with the residual sensitivity as shading. Bottom, angular density of the samples against the data (dashed) and $\sigma(\theta)$ (grey). The first column uses \texttt{flat}. Right, residual vs. distance to the data plane which measures leakage.}
\label{fig:toy_anchored}
\end{figure*}

Figure~\ref{fig:toy_anchored} and Table~\ref{tab:toy} show the anchored objective on the \texttt{graded} circle at $D=10$. Unlike PIDM, the anchored samples cover the circle uniformly at every weight, and their distance to the data plane stays at the level of DM, three to nine times smaller than that of PIDM. The anchored objective therefore does not lower the residual through avoidance or leakage, yet its residual is still below that of DM at every weight.

\begin{table}[t]
\centering\small
\setlength{\tabcolsep}{5pt}
\caption{Shocked circle, \texttt{graded}, $D=10$. $\overline{|R|}$ measures physical accuracy, WD (circular, on the angle) distributional accuracy, the $\sigma$ ratio avoidance and the distance to the data plane leakage.}
\label{tab:toy}
\begin{tabular}{llrrrr}
\toprule
 & $c$ & $\overline{|R|}$~$\downarrow$ & WD~$\downarrow$ & $\sigma$ ratio~$\to1$ & $\|x_{3:D}\|$~$\downarrow$ \\
\midrule
DM & -- & 9.12e-3 & 8.66e-3 & 0.978 & 3.23e-4 \\
\midrule
PIDM & $10^{-3}$ & 2.53e-3 & 2.38e-2 & 0.582 & 1.38e-3 \\
 & $10^{-2}$ & 1.90e-3 & 2.91e-2 & 0.356 & 8.60e-4 \\
 & $10^{-1}$ & 1.04e-3 & 4.15e-2 & 0.231 & 1.36e-3 \\
 & $1$ & 5.15e-4 & 7.80e-2 & 0.201 & 3.30e-3 \\
\midrule
Anchored & $10^{-3}$ & 7.67e-3 & 1.17e-2 & 0.973 & 1.49e-4 \\
 & $10^{-2}$ & 5.70e-3 & 9.93e-3 & 0.980 & 2.60e-4 \\
 & $10^{-1}$ & 4.19e-3 & 7.79e-3 & 0.980 & 2.68e-4 \\
 & $1$ & 3.52e-3 & 9.29e-3 & 0.990 & 4.03e-4 \\
\bottomrule
\end{tabular}
\end{table}

\FloatBarrier
\subsection{Burgers}
\label{app:burgers_detail}

The PIDM results fall into three
regimes (Table~\ref{tab:burgers}). At $c=10^{-7}$ and $10^{-6}$ the physics term works as intended. The residual halves from $0.25$ for DM to $0.11$, at the price of smoother fields, with a
$\mathrm{std}(\sigma)$ ratio of $0.65$ against $0.82$ and an MMD of $0.57$ against $0.12$. From
$c=10^{-5}$ to $10^{-2}$ the fields keep getting smoother and their amplitude shrinks, to a $\mathrm{std}(u)$ ratio of $0.075$, but
the residual no longer falls and varies between $0.15$ and $0.41$. The samples are then neither
close to the data nor solutions of the PDE. 
A likely explanation is that during training the model only sees noisy versions of real fields, and on these inputs its residual does fall monotonically. During sampling, however, the model predicts an almost constant field from pure noise already in the first step for $c\geq10^{-4}$, so all later steps receive inputs it never saw in training (Fig.~\ref{fig:burgers_trajectories}).
From $c=10^{-1}$ on the residual drops below $0.04$,
but the MMD stays at $1.2$. The model is collapsed and generates weak, smooth fields, which leave a small
residual but no longer resemble the data. A low residual of PIDM is therefore no sign of better
physics here. The $\sigma$ ratio falls with the amplitude, from $0.65$ to below $0.1$, which is
the avoidance predicted in Sec.~\ref{sec:avoidance}.

\begin{figure}[!htbp]
\centering
\includegraphics[width=0.6\linewidth]{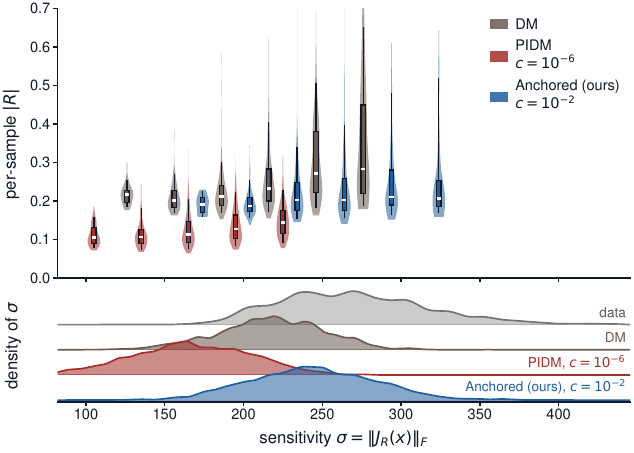}
\caption{Burgers samples by sensitivity $\sigma$. Top, per-sample residual of DM, the anchored objective ($c=10^{-2}$) and PIDM ($c=10^{-1}$, left). Bottom, density of $\sigma$ for the data and each model. PIDM generates only samples far less sensitive than the data, while the anchored samples are closer to the data than those of DM and have a lower residual in every sensitivity range.}

\label{fig:avoid_burger_violin}
\end{figure}

The anchored objective shows none of these regimes. At small weights it even counteracts the tendency of DM toward smooth, insensitive fields, visible in its $\sigma$ ratio of $0.82$. At
$c=10^{-2}$ and $10^{-1}$ the $\sigma$ ratio rises to $0.91$ and $0.89$, the $\mathrm{std}(u)$ ratio to $0.90$ and $0.88$, and the MMD falls to $0.037$ and $0.052$, while the residual still drops below that of DM ($0.23$ against $0.25$). The anchored samples are thus more sensitive than those of DM and still more accurate. Fig.~\ref{fig:avoid_burger_violin} shows why. For DM the median residual grows with $\sigma$ from $0.20$ to $0.28$, since the same error yields a larger residual at a more sensitive state. For the anchored objective it stays between $0.19$ and $0.21$ in every sensitivity range, below that of DM. Larger weights lower the residual to $0.11$, the value PIDM reaches at its smallest weight. Beyond $c=10$ the residual of the anchored objective no longer improves, and its samples become smoother, as the shocks are where the residual is largest. This smoothing is weaker than that of PIDM, and the fields do not collapse. Up to $c=10^{4}$ the $\mathrm{std}(u)$ ratio stays between $0.63$ and $0.72$, while PIDM drops to $0.16$ at $c=10^{-4}$ and below $0.1$ at larger weights. On Burgers anchoring thus avoids the shortcut, and only at large weights does it trade some sharpness for no further gain in residual.

\begin{table}[t]
\centering\small
\setlength{\tabcolsep}{3pt}
\caption{Burgers.}
\label{tab:burgers}
\begin{tabular}{@{}lrrrr@{}}
\toprule
$c$ & \shortstack{$\overline{|R|}$\\$\downarrow$} & \shortstack{MMD\\$\downarrow$} & \shortstack{$\sigma$\\ratio~$\to\!1$} & \shortstack{$\mathrm{std}(u)$\\ratio~$\to\!1$} \\
\midrule
DM & 0.25 & 0.12 & 0.82 & 0.81 \\
\midrule
\multicolumn{5}{@{}l}{\textit{PIDM}} \\
\hspace{0.6em}$10^{-7}$ & 0.11 & 0.57 & 0.65 & 0.64 \\
\hspace{0.6em}$10^{-6}$ & 0.13 & 0.56 & 0.62 & 0.62 \\
\hspace{0.6em}$10^{-5}$ & 0.25 & 0.95 & 0.46 & 0.45 \\
\hspace{0.6em}$10^{-4}$ & 0.22 & 1.2 & 0.13 & 0.16 \\
\hspace{0.6em}$10^{-3}$ & 0.15 & 1.3 & 0.13 & 9.1e-2 \\
\hspace{0.6em}$10^{-2}$ & 0.41 & 1.4 & 7.5e-2 & 7.5e-2 \\
\hspace{0.6em}$10^{-1}$ & 3.1e-2 & 1.2 & 0.31 & 0.33 \\
\hspace{0.6em}$1$ & 9.6e-3 & 1.2 & 9.5e-2 & 8.4e-2 \\
\midrule
\multicolumn{5}{@{}l}{\textit{Anchored}} \\
\hspace{0.6em}$10^{-3}$ & 0.23 & 0.37 & 0.71 & 0.70 \\
\hspace{0.6em}$10^{-2}$ & 0.23 & 3.7e-2 & 0.91 & 0.90 \\
\hspace{0.6em}$10^{-1}$ & 0.23 & 5.2e-2 & 0.89 & 0.88 \\
\hspace{0.6em}$1$ & 0.17 & 0.20 & 0.79 & 0.77 \\
\hspace{0.6em}$10^{1}$ & 0.12 & 0.42 & 0.73 & 0.72 \\
\hspace{0.6em}$10^{2}$ & 0.11 & 0.44 & 0.72 & 0.71 \\
\hspace{0.6em}$10^{3}$ & 0.11 & 0.66 & 0.64 & 0.63 \\
\hspace{0.6em}$10^{4}$ & 0.11 & 0.67 & 0.65 & 0.64 \\
\bottomrule
\end{tabular}
\end{table}

\begin{figure}[!htbp]
\centering
\includegraphics[width=0.6\linewidth]{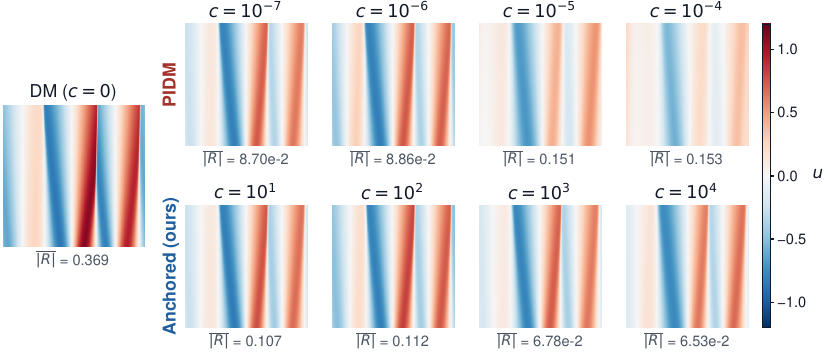}
\caption{Burgers fields generated a standard diffusion model vs. PIDM and anchored with increasing
penalty weights using the same seed.}
\label{fig:avoid_burger_images}
\end{figure}

\begin{figure*}[!htbp]
\centering
\includegraphics[width=0.9\textwidth]{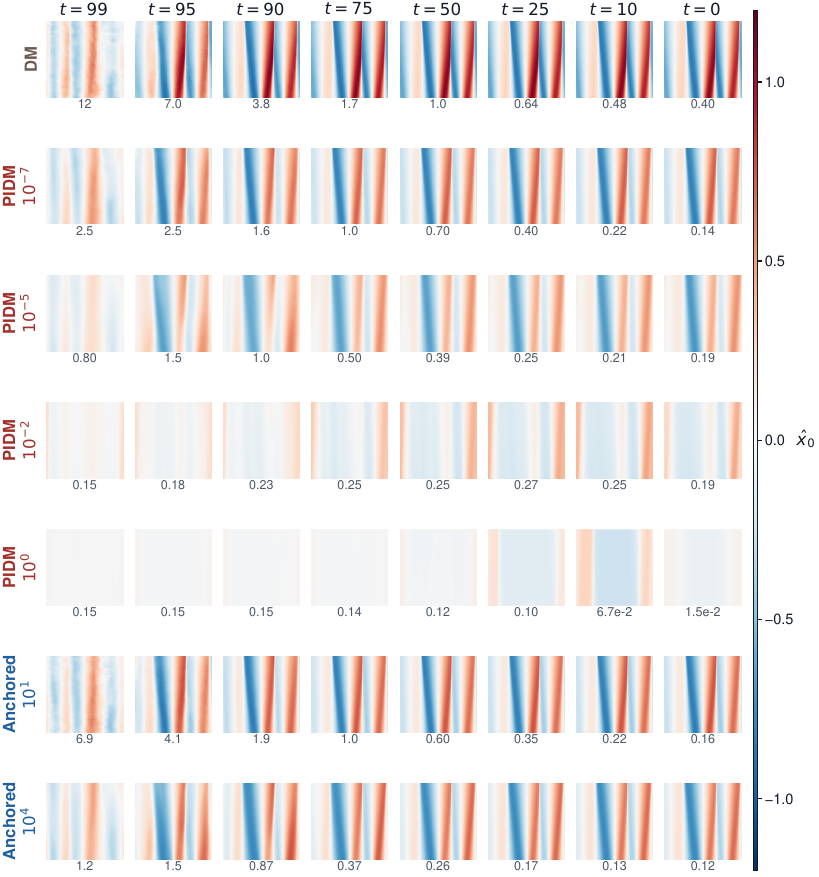}
\caption{Sampling trajectories on Burgers. Each row is one model and each column one step $t$ of
the sampler, from pure noise ($t=99$) to the final sample ($t=0$). Every panel shows the model's
prediction $\hat x_0$ at that step, and the number below it is the mean residual
$\overline{|R|}$ of that prediction. All rows start from the same noise and receive the same noise
in every step, so the last column equals the corresponding fields in
Fig.~\ref{fig:avoid_burger_images}.}

\label{fig:burgers_trajectories}
\end{figure*}

\FloatBarrier
\subsection{Darcy}
\label{app:darcy_detail}
All detailed results are listed in Table~\ref{tab:darcy}.  Up to $c=10^{-5}$ PIDM lowers the residual from $6.0$ for DM to $0.28$, with an MMD at the level of DM ($0.046$ against $0.054$, Table~\ref{tab:darcy}). At $c=10^{-4}$ the pressure flattens, with a $\mathrm{std}(p)$ ratio of $0.64$, the MMD rises to $0.20$ and the residual stops falling ($0.33$). This is the failure predicted in Sec.~\ref{sec:avoidance}. At the comparable weight $10^{6}$ the anchored objective keeps the pressure, with a $\mathrm{std}(p)$ ratio of $0.92$ and an MMD of $0.079$, and reaches a lower residual of $0.24$. Its run at $c=10^{5}$ has the lowest MMD of all runs ($0.028$) and a $\mathrm{std}(p)$ ratio of $0.99$. At the largest weights both objectives flatten the pressure ($\mathrm{std}(p)$ ratio $0.14$ and $0.31$). 

\begin{figure}[!htbp]
\centering
\includegraphics[width=0.6\linewidth]{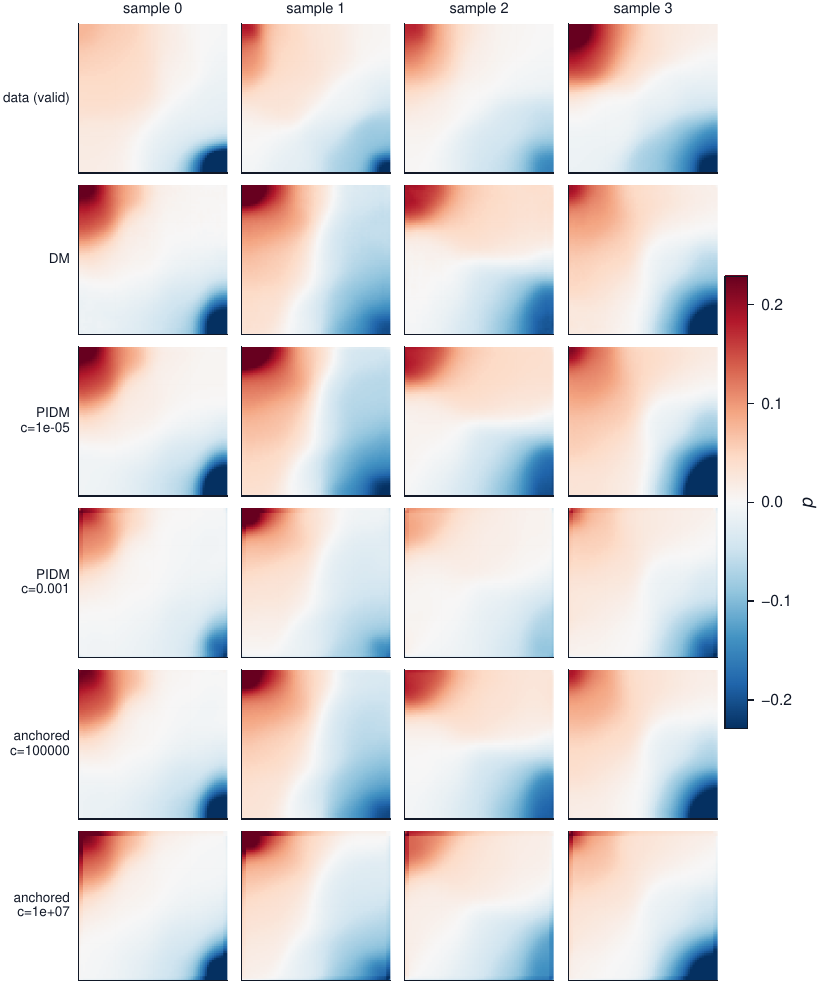}
\caption{Darcy pressure fields generated from the same starting noise by DM, PIDM and the
anchored objective at two weights each.}
\label{fig:darcy}
\end{figure}

Fig.~\ref{fig:darcy} shows generated pressure fields from the same starting noise. With
increasing weight the PIDM pressure flattens, while the anchored samples keep the variation of
the data at every weight.

\begin{table}[t]
\centering\small
\setlength{\tabcolsep}{3pt}
\caption{Darcy. Rows of PIDM and the anchored objective with comparable weights are aligned. The anchored objective reaches a
lower residual at comparable MMD.}
\label{tab:darcy}
\begin{tabular}{@{}lrrrrrr@{}}
\toprule
$c$ & \shortstack{$\overline{|R|}$\\$\downarrow$} & \shortstack{forward\\error\\$\downarrow$} & \shortstack{MMD\\$\downarrow$} & \shortstack{$\sigma$\\ratio\\$\to\!1$} & \shortstack{$\mathrm{std}(p)$\\ratio\\$\to\!1$} & \shortstack{$\mathrm{std}(K)$\\ratio\\$\to\!1$} \\
\midrule
DM & 6.0 & 3.4e-2 & 5.4e-2 & 0.84 & 1.1 & 0.75 \\
\midrule
\multicolumn{7}{@{}l}{\textit{PIDM}} \\
\hspace{0.6em}$10^{-7}$ & 1.2 & 4.3e-2 & 7.3e-2 & 0.80 & 1.2 & 0.72 \\
\hspace{0.6em}$10^{-6}$ & 0.96 & 5.5e-2 & 4.7e-2 & 0.85 & 1.1 & 0.74 \\
\hspace{0.6em}$10^{-5}$ & 0.28 & 0.19 & 4.6e-2 & 0.83 & 0.98 & 0.73 \\
\hspace{0.6em}$10^{-4}$ & 0.33 & 0.39 & 0.20 & 0.91 & 0.64 & 0.75 \\
\hspace{0.6em}$10^{-3}$ & 0.14 & 0.87 & 0.66 & 0.79 & 0.14 & 0.66 \\
\midrule
\multicolumn{7}{@{}l}{\textit{Anchored}} \\
\hspace{0.6em}$10^{3}$ & 2.0 & 5.9e-2 & 5.3e-2 & 0.84 & 1.1 & 0.75 \\
\hspace{0.6em}$10^{4}$ & 0.56 & 3.6e-2 & 5.8e-2 & 0.83 & 1.1 & 0.74 \\
\hspace{0.6em}$10^{5}$ & 0.35 & 4.7e-2 & 2.8e-2 & 0.90 & 0.99 & 0.80 \\
\hspace{0.6em}$10^{6}$ & 0.24 & 0.22 & 7.9e-2 & 0.82 & 0.92 & 0.71 \\
\hspace{0.6em}$10^{7}$ & 0.36 & 1.0 & 0.90 & 0.79 & 0.31 & 0.79 \\
\bottomrule
\end{tabular}

\end{table}

\FloatBarrier
\subsection{Advection}
\label{app:advection_detail}

We pair a PIDM weight $c$ with the anchored weight $2\cdot10^{6}c$. On Advection the residual is
linear, so the anchored term equals the PIDM term divided by the constant
$\|J_R\|_F^2/m=6.05\cdot10^{4}$, and the remaining factor equalizes the average of the two
noise-level weightings (App.~\ref{app:implementation}).

\begin{table}[!htbp]
\centering\small
\setlength{\tabcolsep}{3pt}
\caption{Advection. PIDM (left) and the anchored objective (right) with comparable weights in the same row. DM reaches $\overline{|R|}=1.6\text{e-}2$, an MMD of $0.28$ and a $\mathrm{std}(u)$ ratio of $0.51$.}
\label{tab:advection}
\begin{tabular}{@{}lrrr@{\hspace{1.2em}}lrrr@{}}
\toprule
\multicolumn{4}{@{}c}{\textit{PIDM}} & \multicolumn{4}{c@{}}{\textit{Anchored}} \\
\cmidrule(r{1.2em}){1-4}\cmidrule{5-8}
$c$ & \shortstack{$\overline{|R|}$\\$\downarrow$} & \shortstack{MMD\\$\downarrow$} & \shortstack{$\mathrm{std}(u)$\\ratio~$\to\!1$}
& $c$ & \shortstack{$\overline{|R|}$\\$\downarrow$} & \shortstack{MMD\\$\downarrow$} & \shortstack{$\mathrm{std}(u)$\\ratio~$\to\!1$} \\
\midrule
$10^{-3}$ & 4.6e-3 & 0.45 & 0.37    & $2\cdot10^{3}$ & 5.3e-3 & 0.29 & 0.50 \\
$10^{-2}$ & 2.2e-3 & 0.59 & 0.24    & $2\cdot10^{4}$ & 4.2e-3 & 0.37 & 0.41 \\
$10^{-1}$ & 1.3e-3 & 0.84 & 5.6e-2  & $2\cdot10^{5}$ & 1.6e-3 & 0.61 & 0.17 \\
$1$       & 9.2e-4 & 0.87 & 7.2e-3  & $2\cdot10^{6}$ & 9.1e-4 & 0.87 & 4.1e-3 \\
\bottomrule
\end{tabular}
\end{table}

\end{document}